\documentclass[10pt,twocolumn,letterpaper]{article}

\usepackage[cvprfinal]{cvpr}   
\usepackage{bigfoot}

\usepackage{mathrsfs}
\usepackage{graphicx}
\usepackage{booktabs}
\usepackage{multirow}
\usepackage{makecell}
\usepackage{xcolor}
\usepackage{colortbl}
\usepackage{lipsum}
\usepackage{pgffor}
\usepackage{xcolor}
\usepackage{svg}

\newcommand\blfootnote[1]{%
  \begingroup
  \renewcommand\thefootnote{}\footnote{#1}%
  \addtocounter{footnote}{-1}%
  \endgroup
}

\definecolor{colorfirst}{rgb}{.866,.945, 0.831} 
\definecolor{colorsecond}{rgb}{1, 0.98, 0.83} 
\definecolor{colorthird}{rgb}{0.76, 0.87, 0.92} 
\definecolor{colorcite}{rgb}{0.212, 0.490, 0.741} 

\newcommand{\cellfirst}{\cellcolor{colorfirst}}
\newcommand{\cellsecond}{\cellcolor{colorsecond}}
\newcommand{\cellthird}{\cellcolor{colorthird}}

\newcommand{\textfirst}{\colorbox{colorfirst}}
\newcommand{\secondtext}{\colorbox{colorsecond}}
\newcommand{\thirdtext}{\colorbox{colorthird}}

\newcommand{\citecolor}[1]{{\color{colorcite}#1}}

\newcommand{\fullname}{{\textit{Prompt Depth Anything}}\xspace}

\newcommand{\symlidar}{\mathbf{L}}

\renewcommand{\paragraph}[1]{\vspace{.3em}\noindent\textbf{#1}}

\definecolor{cvprblue}{rgb}{0.21,0.49,0.74}
\usepackage[pagebackref,breaklinks,colorlinks,allcolors=cvprblue]{hyperref}

\def\paperID{68888} 
\def\confYear{2070}

\title{Prompting Depth Anything for 4K Resolution Accurate Metric Depth Estimation}

\author{
\\[-20pt]
Haotong Lin\textsuperscript{1,2} \quad
Sida Peng\textsuperscript{1$^\dagger$} \quad
Jingxiao Chen\textsuperscript{3} \quad 
Songyou Peng\textsuperscript{4} \quad
Jiaming Sun\textsuperscript{1} \\[1pt]
Minghuan Liu\textsuperscript{3} \quad
Hujun Bao\textsuperscript{1} \quad
Jiashi Feng\textsuperscript{2} \quad
Xiaowei Zhou\textsuperscript{1} \quad
Bingyi Kang\textsuperscript{2}
\\[6pt]
    $^1$Zhejiang University \quad
    $^2$ByteDance Seed \quad
    $^3$Shanghai Jiao Tong University \quad
    $^4$ETH Zurich \\[-10pt]
\\[-1pt]
{\normalsize \url{https://PromptDA.github.io/}}
}

\begin{document}

\twocolumn[\maketitle\vspace{0em}\begin{center}
    \vspace{-8mm}
    \includegraphics[width=\linewidth]{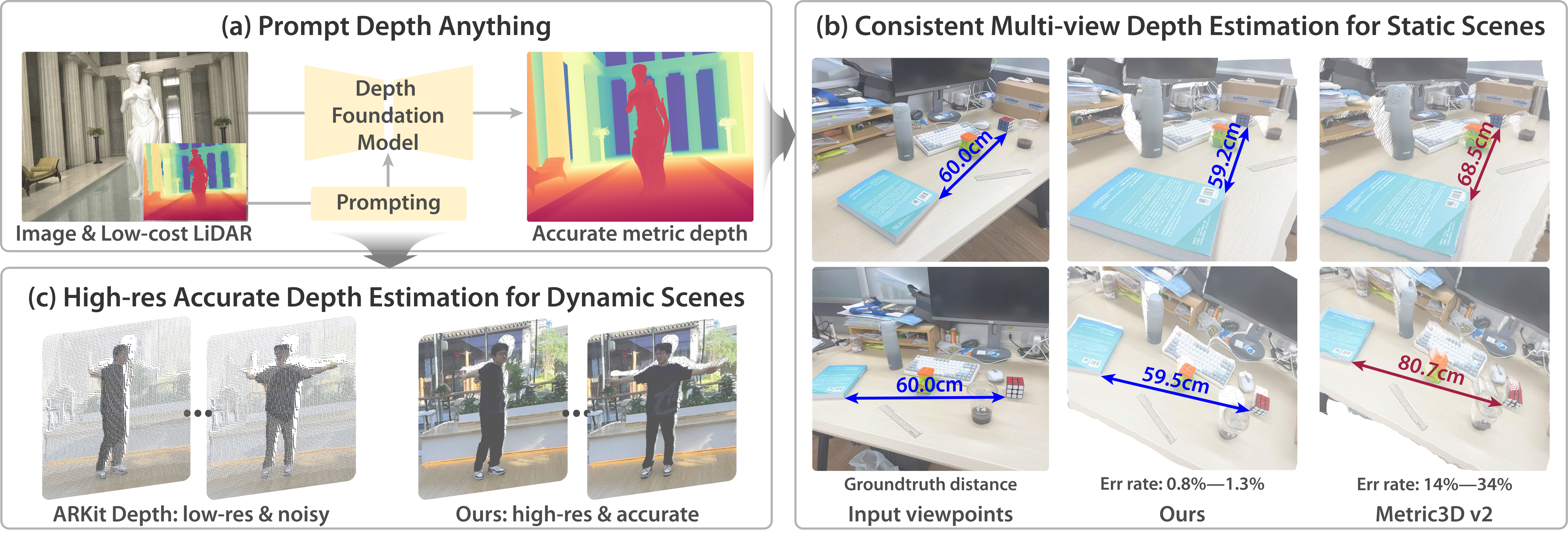}
    \vspace{-8mm}
\end{center}
\captionof{figure}{
\textbf{Illustration and capabilities of \textit{Prompt Depth Anything}}.
 (a) \fullname is a new paradigm for metric depth estimation, which is formulated as prompting a depth foundation model with a metric prompt, specifically utilizing a low-cost LiDAR as the prompt. 
 (b) Our method enables consistent depth estimation, addressing the limitations of Metric3D v2~\cite{hu2024metric3dv2} that suffer from inaccurate scale and inconsistency.
 (c) It achieves accurate 4K accurate depth estimation, 
 significantly surpassing ARKit LiDAR Depth (240 $\times$ 320). 
}
\label{fig:teaser}
\bigbreak]

\blfootnote{$^\dagger$Corresponding author: Sida Peng}

\begin{abstract}
Prompts play a critical role in unleashing the power of language and vision foundation models for specific tasks. For the first time, we introduce prompting into depth foundation models, creating a new paradigm for metric depth estimation termed \textbf{Prompt Depth Anything}. Specifically, we use a low-cost LiDAR as the prompt to guide the Depth Anything model for accurate metric depth output, achieving up to 4K resolution. 
Our approach centers on a concise prompt fusion design that integrates the LiDAR at multiple scales within the depth decoder. 
To address training challenges posed by limited datasets containing both LiDAR depth and precise GT depth, we propose a scalable data pipeline that includes synthetic data LiDAR simulation and real data pseudo GT depth generation. 
Our approach sets new state-of-the-arts on the ARKitScenes and ScanNet++ datasets 
and benefits downstream applications, including 3D reconstruction and generalized robotic grasping.
\end{abstract}
    
\section{Introduction}
\label{sec:intro}
High-quality depth perception is a fundamental challenge in computer vision and robotics.
Recent monocular depth estimation has experienced a significant leap by scaling the model or data, leading to the flourishing of depth foundation models~\cite{yang2024depth,yang2024depthv2,ke2024repurposing,fu2025geowizard}. 
These models demonstrate strong abilities in producing high-quality relative depth, but suffer from scale ambiguity, hindering their practical applications in autonomous driving and robotic manipulation, \etc.
Therefore, significant efforts have been made to achieve metric depth estimation, by either finetuning depth foundation models~\cite{bhat2023zoedepth,guizilini2023towards} on metric datasets or training metric depth models with image intrinsics as additional inputs~\cite{piccinelli2024unidepth,yin2023metric3d,bochkovskii2024depth,hu2024metric3dv2}.
However, neither of them can address the problem properly, as illustrated in \cref{fig:teaser}\citecolor{(b)}.

A natural question thus arises: \emph{Do these foundation models truly lack utility in accurate metric depth estimation}? This reminds us to closely examine the foundation models in natural language~\cite{brown2020language,achiam2023gpt} and vision~\cite{rombach2021highresolution,liu2023llava,liu2023improvedllava}, which often involve pre-training and instruction tuning stages. A properly designed \textbf{\textit{prompt}} and a \textbf{\textit{instruction dataset}} can unlock the power of foundation models on downstream tasks. 
Inspired by these successes, we propose a new paradigm for metric depth estimation by treating it as a downstream task, \textit{i.e.}, prompting a depth foundation model with metric information. We believe this prompt can take any form as long as the scale information is provided, \eg, camera intrinsics. In this paper, we validate the feasibility of the paradigm by choosing low-cost LiDAR as the prompt for two reasons. First, it provides precise metric scale information. Second, it is widely available, even in common mobile devices (\eg, Apple iPhone has a LiDAR).



Specifically, based on Depth Anything~\cite{yang2024depthv2}, we propose \textit{\textbf{\fullname}}, which achieves 4K resolution accurate metric depth estimation.
At the core of our method is a concise prompt fusion architecture tailored for the DPT-based~\cite{ranftl2021vision} depth foundation models~\cite{yang2024depthv2,bochkovskii2024depth}.
The prompt fusion architecture integrates the LiDAR depth at multiple scales within the DPT decoder, fusing the LiDAR features for depth decoding.
The metric prompt provides precise spatial distance information, making the depth foundation model particularly serve as a local shape learner, resulting in accurate and high-resolution metric depth estimation.

Training \fullname requires both LiDAR depth and precise GT depth. However, existing synthetic data~\cite{roberts2021} lacks LiDAR depth, and real-world data~\cite{yeshwanth2023scannet++} with LiDAR only has an imprecise GT depth of bad edges. 
To solve this challenge, we propose a scalable data pipeline that simulates low-resolution, noisy LiDAR for synthetic data and generates pseudo GT depth with high-quality edges for real data using a reconstruction method~\cite{barron2023zip}. 
To mitigate errors in the pseudo GT depth from the 3D reconstruction, we introduce an edge-aware depth loss that leverages only the gradient of pseudo GT depth, which is prominent at edges. We experimentally demonstrate that these efforts result in highly accurate depth estimation.

We evaluate the proposed method on ARKitScenes~\cite{baruch2021arkitscenes} and ScanNet++~\cite{yeshwanth2023scannet++} datasets containing iPhone ARKit depth. It consistently exhibits state-of-the-art performance across datasets and metrics.
Even our zero-shot model achieves better performance compared to other methods~\cite{yang2024depthv2,bhat2023zoedepth} in non-zero-shot testing, highlighting the generalization ability of prompting a foundation model.
We also show that the foundation model and prompt of \fullname can be replaced with DepthPro~\cite{bochkovskii2024depth} and vehicle LiDAR~\cite{sun2020scalability}, respectively.
Furthermore, we demonstrate that it benefits several downstream applications, including 3D reconstruction and generalized robotic object grasping.

In summary, this work has the following contributions:
\begin{itemize}
    \item \textit{Prompt Depth Anything}, a new paradigm for metric depth estimation by prompting a depth foundation model with a low-cost LiDAR as the metric prompt.
    \item A concise prompt fusion architecture for depth foundation models, a scalable data pipeline, and an edge-aware depth loss to train \textit{Prompt Depth Anything}.
    \item State-of-the-art performance on depth estimation benchmarks~\cite{yeshwanth2023scannet++,baruch2021arkitscenes}, showing the extensibility of replacing depth foundation models and LiDAR sensors, and highlighting benefits for several downstream applications including 3D reconstruction and robotic object grasping.
\end{itemize}

\section{Related Work}

\paragraph{Monocular depth estimation.}
Traditional methods~\cite{saxena2008make3d,hoiem2007recovering} rely on hand-crafted features for depth estimation. With the advent of deep learning, this field has seen significant advancements. Early learning-based approaches~\cite{eigen2014depth,eigen2015predicting} are often limited to a single dataset, lacking generalization capabilities. To enhance generalization, diverse datasets~\cite{Li2018:CVPR,blendedmvs,diml,hrwsi,tartanair,irs,wsvd,redweb}, affine-invariant loss~\cite{midas}, and more powerful network architectures~\cite{ranftl2021vision} have been introduced. More recently, latent diffusion models~\cite{rombach2021highresolution}, pre-trained on extensive image generation tasks, have been applied to depth estimation~\cite{ke2024repurposing,he2024lotus}. These models exhibit good generalization, estimating relative depth effectively, though they remain scale-agnostic. To achieve metric depth estimation, early methods either model the problem as global distribution classification~\cite{Fu2018:CVPR,Bhat2021:CVPR,Bhat2022:ECCV,Li2024:TIP} or fine-tune a depth model on metric depth datasets~\cite{bhat2023zoedepth,li2023patchfusion,li2024patchrefiner}. Recent methods~\cite{yin2023metric3d,guizilini2023towards,yin2021learning,hu2024metric3dv2,piccinelli2024unidepth} discuss the ambiguity in monocular metric depth estimation and address it by incorporating camera intrinsic parameters.
Although recent methods~\cite{yin2023metric3d,hu2024metric3dv2,piccinelli2024unidepth,bochkovskii2024depth,yang2024depthv2,ke2024repurposing,he2024lotus} exhibit strong generalization ability and claim to be depth foundation models~\cite{bochkovskii2024depth,yang2024depthv2,hu2024metric3dv2,fu2025geowizard}, metric depth estimation remains a challenge as shown in \cref{fig:teaser}\citecolor{(b)}. We seek to address this challenge by prompting the depth foundation models with a metric prompt, inspired by the success of prompting in vision and vision-language models~\cite{liu2023llava,liu2023improvedllava,zhang2023adding}. 

\begin{figure*}[t]
    \centering 
    \includegraphics[width=0.99\textwidth]{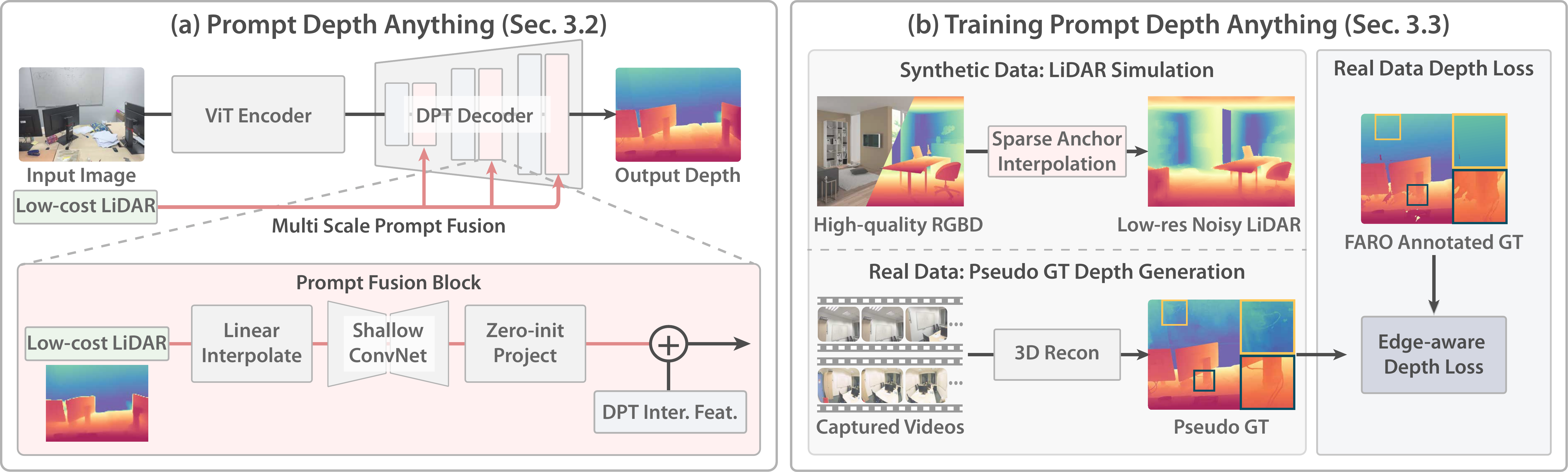}
    \vspace{-3pt}
    \caption{\textbf{Overview of \fullname.}
    (a)  \fullname builds on a depth foundation model~\cite{yang2024depthv2} with a ViT encoder and a DPT decoder, and adds a multi-scale prompt fusion design, using a prompt fusion block to fuse the metric information at each scale.
    (b) 
    Since training requires both low-cost LiDAR and precise GT depth, we propose a scalable data pipeline that simulates LiDAR depth for synthetic data with precise GT depth, and generates pseudo GT depth for real data with LiDAR.
    An edge-aware depth loss is proposed to merge accurate edges from pseudo GT depth with accurate depth in textureless areas from FARO annotated GT depth on real data.
    }
    \vspace{-8pt}
    \label{fig:pipeline}
\end{figure*}

\paragraph{Depth estimation with auxiliary sensors.}
Obtaining dense depth information through active sensors typically demands high power consumption~\cite{xu2022attention,xu2023accurate,xu2023iterative,xu2024igevplus,xu2024towards,cheng2025monster}. 
A more practical approach involves utilizing a low-power active sensor to capture sparse depth, which can then be completed into dense maps. 
Many studies investigate methods to fill in sparse depth data. 
Early works rely on filter-based~\cite{guide_filter,mean_filter,bilateral_upsampling} and optimization-based~\cite{TV,autoregression} techniques for depth completion. More recent studies~\cite{BP-Net,cformer,cspn++,cspn,guidenet,graphcspn,sun2023consistent,xu2024towards,lin2022dynamic,conti2024depth,guo2025murre} adopt learning-based approaches for depth completion.
Typically, these methods are not tested on real indoor LiDAR data but rather on simulated sparse lidar for depth datasets such as NYUv2~\cite{eigen2014depth} to reconstruct complete depth. This is because real testing setups require both low-power and high-power LiDAR sensors. 
More recent works have collected both low-power and high-power LiDAR data. 
To collect such data, DELTA~\cite{li2022deltar} builds a suite to collect data using L5 and Intel RealSense 435i, while three other datasets~\cite{baruch2021arkitscenes,yeshwanth2023scannet++,ren2024mushroom} are collected using iPhone LiDAR and FARO LiDAR. We focus on the latter, as iPhone is widely available.
A recent work similar to ours is Depth Prompting~\cite{park2024depth}.
Our approach differs in that we use a network to take sparse depth as a prompt for the depth foundation model, achieving specific output. In contrast, they fuse sparse depth with features from the depth foundation model to post-process the foundation model output, which does not constitute prompting a foundation model.



\section{Method}
\label{sec:method}
Monocular depth estimation models~\cite{yang2024depth,yang2024depthv2,bochkovskii2024depth} are becoming depth foundation models for their generalization ability obtained from large-scale data. 
However, due to the inherent ambiguities, they cannot achieve high accuracy on metric depth estimation as shown in~\cref{fig:teaser}\citecolor{(b)}.
Inspired by the success of prompting for vision~\cite{kirillov2023segment,rombach2021highresolution,liu2023llava} and language~\cite{achiam2023gpt} foundation models, we propose \fullname prompting the depth foundation model with a metric prompt to achieve metric depth estimation.
We take the low-cost LiDAR as the metric prompt in this work, as it has recently been integrated into lots of smartphones, making this setup highly practical.
%
To be specific, we aim to prompt the depth foundation model to unleash its power for accurate metric depth estimation.

\subsection{Preliminary: Depth Foundation Model}
\label{sec:method:pre}
Current depth foundation models~\cite{yang2024depth,yang2024depthv2,yin2023metric3d,birkl2023midas} generally share similar network structures of DPT~\cite{ranftl2021vision} networks.
Specifically, given an image $\mathbf{I} \in \mathbb{R}^{C \times H \times W}$, they take a vision transformer (ViT) with multiple stages to extract tokenized image features $\{\mathbf{T}_i\}$, where $\mathbf{T}_i \in \mathbb{R}^{C_i \times (\frac{H}{p} \times \frac{W}{p} + 1)}$ represents the feature map at stage $S_i$, $D_i$ is the feature dimension at stage $S_i$, and $p$ is the patch size.
The DPT decoder reassembles features from different stages into image-like representations $\mathbf{F}_{i} \in \mathbb{R}^{D_i \times \frac{H}{p} \times \frac{W}{p}}$ with the reassemble operation~\cite{ranftl2021vision}.
Finally, a sequence of convolutional blending steps are applied to merge features ${\mathbf{F}_i}$ across different stages, predicting a dense depth map $\mathbf{D} \in \mathbb{R}^{H \times W}$. 

We note that there exists another line of depth foundation models~\cite{ke2024repurposing,fu2025geowizard,he2024lotus} that use the image diffusion model~\cite{rombach2022high} to estimate depth maps.
Due to the high computational cost of diffusion models, we only consider DPT-based depth foundation models~\cite{yang2024depthv2,bochkovskii2024depth} as our base model for real-time performance in this work.

\subsection{Prompt Depth Anything}
\label{sec:method:fusion}
In this section, we seek to find a concise way to incorporate a low-cost LiDAR (i.e., a low-resolution and noisy depth map) as a prompt into the depth foundation model.
To this end, we propose a concise prompt fusion architecture tailored for the DPT-based~\cite{ranftl2021vision} depth foundation models to integrate low-resolution depth information.
As shown in \cref{fig:pipeline}\citecolor{(a)}, the prompt fusion architecture integrates low-resolution depth information at multiple scales within the DPT Decoder.
Specifically, for each scale $S_i$ in the DPT Decoder, a low-resolution depth map $\symlidar \in \mathbb{R}^{1 \times H_{\symlidar} \times W_{\symlidar}}$ is firstly bilinearly resized to match the spatial dimensions of the current scale $\mathbb{R}^{1 \times H_i \times W_i}$. Then, the resized depth map is passed through a shallow convolutional network to extract depth features. After that, the extracted features are projected to the same dimension as the image features $\mathbf{F}_i \in \mathbb{R}^{C_i \times H_i \times W_i}$ using a zero-initialized convolutional layer. Finally, the depth features are added to the DPT intermediate features for depth decoding.
The illustration of this block design is shown in \cref{fig:pipeline}.


The proposed design has the following advantages. 
Firstly, it introduces only 5.7\% additional computational overhead (1.789 TFLOPs v.s. 1.691 TFLOPs for a $756\times1008$ image) to the original depth foundation model, and effectively addresses the ambiguity issue inherent in the depth foundation model as demonstrated in \cref{tab:ablation}\citecolor{(b)}. 
Secondly, it fully inherits the capabilities of the depth foundation model because its encoder and decoder are initialized from the foundation model~\cite{yang2024depthv2}, and the proposed fusion architecture is zero-initialized, ensuring that the initial output is identical to that of the foundation model.
We experimentally verify the importance of inheriting from a pretrained depth foundation model as shown in \cref{tab:ablation}\citecolor{(c)}.

\paragraph{Optional designs.}
Inspired by conditional image generation methods~\cite{zhang2023adding,peebles2023scalable,karras2019style}, we also explore various potential prompt conditioning designs into the depth foundation model.
Specifically, we experimented with the following designs: a) Adaptive LayerNorm~\cite{perez2018film,karras2019style} which adapts the layer normalization parameters of the encoder blocks based on the conditioning input, b) CrossAttention~\cite{vaswani2017attention} which injects a cross attention block after each self-attention block and integrates the conditioning input through cross-attention mechanisms, and 
c) ControlNet~\cite{zhang2023adding} which copies the encoder blocks and inputs control signals to the copied blocks to control the output depth.
As shown in \cref{tab:ablation}\citecolor{(d,e,f)}, our experiments reveal that these designs do not perform as well as the proposed fusion block. 
A plausible reason is that they are designed to integrate cross-modal information (e.g., text prompts), which does not effectively utilize the pixel alignment characteristics between the input low-res LiDAR and the output depth. 
We detail these optional designs in the supp.

\begin{figure}[t]
    \centering 
    \includegraphics[width=1\columnwidth]{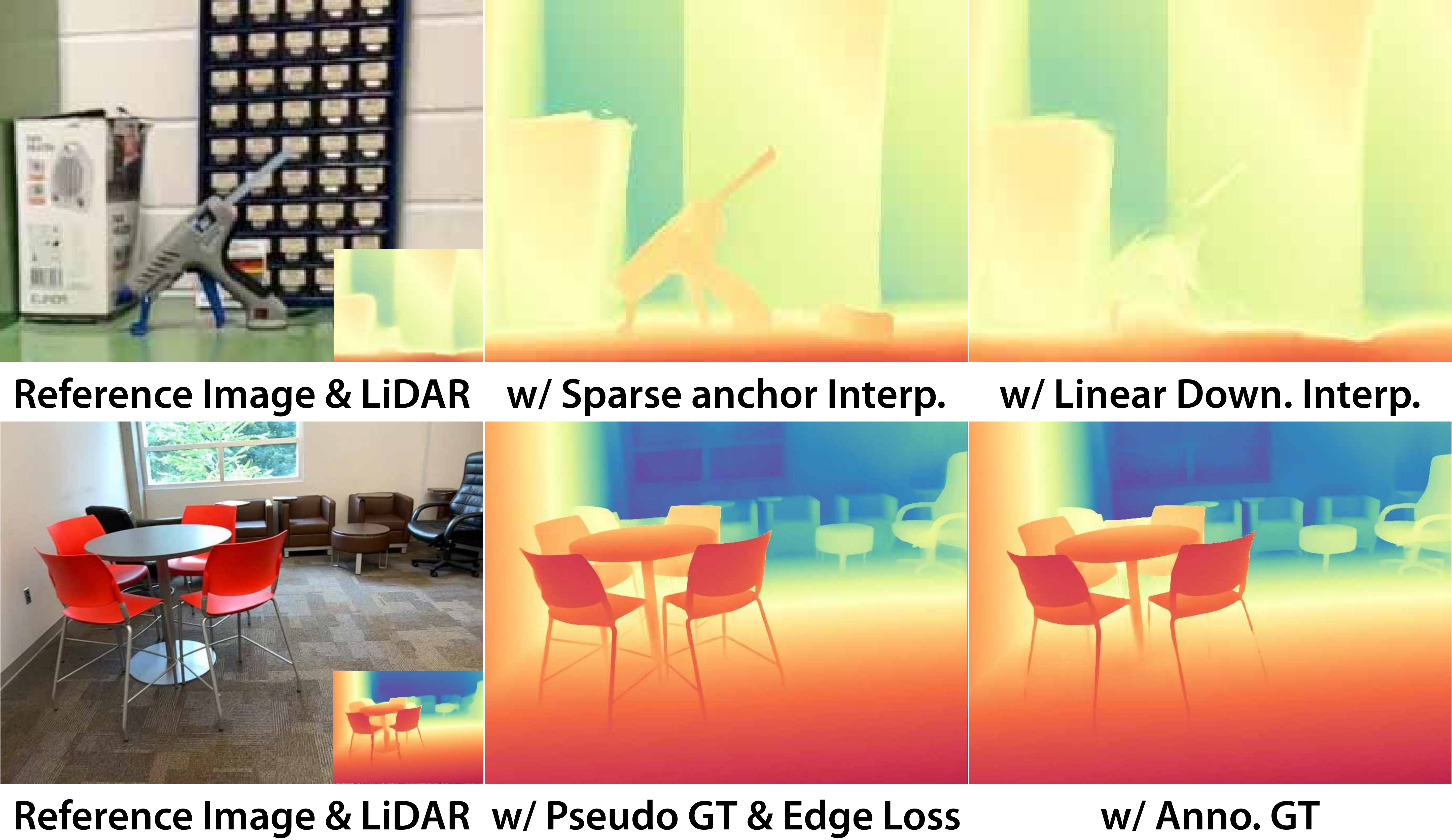}
    \caption{\textbf{Effects on the synthetic data lidar simulation and real data pseudo GT generation with the edge-aware depth loss.} The middle and right columns are the depth prediction results of our different models. The two rows highlight the significance of sparse anchor interpolation for lidar simulation and pseudo GT generation with edge-aware depth loss, respectively.
    }
    \label{fig:ablation}
\end{figure}

\subsection{Training Prompt Depth Anything}
\label{sec:method:data}
Training \fullname simultaneously requires a low-cost LiDAR and precise GT depth.
However, synthetic data~\cite{roberts2021} do not contain LiDAR depth, real-world data with noisy LiDAR depth~\cite{yeshwanth2023scannet++} only have imprecise depth annotations.
Therefore, we propose a LiDAR simulation method for synthetic data and generate pseudo GT depth from ZipNeRF~\cite{barron2023zip} with an edge-aware depth loss for real data. Note that more effective approaches~\cite{xie2024envgs,yan2024street} can be applied.

\paragraph{Synthetic data: LiDAR simulation.}
A LiDAR depth map is low-resolution and noisy. The naive approach for simulating it is to directly downsample the synthetic data depth map. However, this method leads to the model learning depth super-resolution, as shown in \cref{fig:ablation}, meaning that the model does not correct the LiDAR noise. To simulate the noise, we introduce a sparse anchor interpolation method. Specifically, we first downsample the GT depth map to low-resolution ($192 \times 256$, exactly the depth resolution of iPhone ARKit Depth). Then we sample points on this depth map using a distorted grid with a stride (7 in practice). The remaining depth values are interpolated from these points using RGB similarity with KNN. As shown in \cref{fig:ablation}, it effectively simulates LiDAR noise and results in better depth prediction. We provide visualization results of the simulated LiDAR in the supp.




\paragraph{Real Data: Pseudo GT depth generation.} 
We also add real data~\cite{yeshwanth2023scannet++} to our training data.
The annotated depth in ScanNet++~\cite{yeshwanth2023scannet++} is re-rendered from a mesh scanned by a high-power LiDAR sensor (FARO Focus Premium laser scanner). Due to the presence of many occlusions in the scene, several scan positions (typically 4 in a medium-sized scene in ScanNet++) result in an incomplete scanned mesh, leading to depth maps with numerous holes and poor edge quality, as illustrated in \cref{fig:pipeline}\citecolor{(b)}. Motivated by the success of reconstruction methods~\cite{barron2023zip,mildenhall2020nerf}, we propose using Zip-NeRF~\cite{barron2023zip} to recover high-quality depth maps. 
Specifically, we train Zip-NeRF for each scene in ScanNet++ and re-rendered pseudo GT depth.
To provide Zip-NeRF with high-quality and dense observations, 
we detect unblurred frames in Scannet++iPhone videos, and additionally utilize DSLR videos to provide high-quality dense-view images.


\paragraph{Real Data: Edge-aware depth loss.}
Although Zip-NeRF can generate high-quality edge depth, reconstructing textureless and reflective regions remains challenging as shown in \cref{fig:pipeline}\citecolor{(b)}. 
In contrast, these areas (e.g., walls, floors, and ceilings etc.) are usually planar with few occlusions, and the annotations depth in FARO rendered depth is good in these regions. This motivates us to leverage the strengths of both. 
We propose an edge-aware depth loss to meet these requirements. 
Specifically, we use the FARO scanned mesh depth and the gradient of the pseudo GT depth to supervise output depth and the gradient of the output depth, respectively:
\begin{equation}
    \mathcal{L}_{\text{edge}} = L_1(\mathbf{D}_{\text{gt}}, \hat{\mathbf{D}}) + \lambda \cdot \mathcal{L}_{\text{grad}}(\mathbf{D}_{\text{pseudo}}, \hat{\mathbf{D}}),
\end{equation}
\begin{equation}
\mathcal{L}_{\text{grad}}(\mathbf{D}_{\text{pseudo}}, \hat{\mathbf{D}}) = (| \tfrac{\partial (\hat{\mathbf{D}} - \mathbf{D}_{\text{pseudo}})}{\partial x} | + | \tfrac{\partial (\hat{\mathbf{D}} - \mathbf{D}_{\text{pseudo}})}{\partial y} | ).
\end{equation}
In practice, we set $\lambda = 0.5$. 
The depth gradient is mainly prominent at the edges, which is exactly where the pseudo GT depth excels. 
The gradient loss encourages the model to learn the accurate edges from the pseudo GT depth, while the L1 loss encourages the model to learn the overall depth, ultimately leading to excellent depth prediction.
We experimentally verify the effectiveness of the edge-aware depth loss in \cref{tab:ablation}\citecolor{(j)} and \cref{fig:ablation}.

\subsection{Implementation Details}

In this section, we provide essential information about the network design, depth normalization, and training details. Please refer to the supp. for more details.

\paragraph{Network details.}
We utilize the ViT-large model as our backbone model. 
The shallow convolutional network comprises two convolutional layers with a kernel size of 3 and a stride of 1. 
More details can be found in the supp.
Detailed running time analysis can be found in \cref{sec:ablation}.

\paragraph{Depth normalization.}
The irregular range of input depth data can hinder network convergence. To address this, we normalize the LiDAR data using linear scaling to the range [0, 1], based on its minimum and maximum values. The network output is also normalized with the same scaling factor from LiDAR data, ensuring consistent scales and facilitating easier convergence during training. 

\paragraph{Training details.}
We initiate training from the metric model released by Depth Anything v2~\cite{yang2024depthv2}, incorporating a 10K step warm-up phase. During this warm-up phase, we fine-tune this metric model to output a normalized depth derived from the linear scaling of LiDAR data.
Subsequently, we train our model for 200K steps. During the training process, the batch size is set to 2, utilizing 8 GPUs. We employ the AdamW optimizer, with a learning rate of 5e-6 for the ViT backbone and 5e-5 for the other parameters.



\begin{table}
  \setlength\tabcolsep{4 pt}
  \begin{center}
  \resizebox{\columnwidth}{!}{
  \begin{tabular}{l|l|cc|cc|cc}
    \Xhline{3\arrayrulewidth}
    \rule{0pt}{10pt} \multirow{2}{*}{\makecell[c]{Zero \\ Shot}}  & \multirow{2}{*}{\makecell[c]{\textfirst{Net.} / \secondtext{Post.}/ \\ \thirdtext{w/o LiDAR}}} & \multicolumn{2}{c|}{$384 \times 512$}  &  \multicolumn{2}{c|}{768x1024} & \multicolumn{2}{c}{1440x1920} \\
     &  & L1 $\downarrow$ & RMSE $\downarrow$ & L1  $\downarrow$ & RMSE $\downarrow$ & L1  $\downarrow$ & RMSE $\downarrow$ \\ 
     \Xhline{3\arrayrulewidth}
     \rule{0pt}{10pt} \multirow{8}{*}{\makecell[c]{No}} &  \textbf{Ours} \cellfirst &  \textbf{0.0135} &  \textbf{0.0326} &  \textbf{0.0132} &  \textbf{0.0315} &  \textbf{0.0138} &  \textbf{0.0316} \\ 
    & MSPF \cellfirst & 0.0153 & 0.0369 & 0.0149 & 0.0362 & 0.0152 & 0.0363 \\
    \cline{2-8} 
    \rule{0pt}{10pt}  & 
    Depth Pro$^*$  \cellsecond   &   0.0437 & 0.0672 & 0.0435 & 0.0665  & 0.0425 & 0.0654 \\
     & DepthAny. v2$^*$ \cellsecond & 0.0464 & 0.0715 & 0.0423 &  0.0660 & 0.0497 & 0.0764 \\
    & ZoeDepth$^*$ \cellsecond & 0.0831  & 0.2873  & 0.0679 & 0.1421 & 0.0529 & 0.0793 \\
    \cline{2-8} 
    \rule{0pt}{10pt} & 
        Depth Pro$^*$  \cellthird & 0.1222 & 0.1424 &  0.1225 & 0.1427 & 0.1244 & 0.1444  \\
    & DepthAny. v2$^*$  \cellthird & 0.0978 & 0.1180 &  0.0771 &  0.0647& 0.0906 & 0.1125 \\
     & ZoeDepth$^*$  \cellthird & 0.2101  & 0.2784 & 0.1780 & 0.2319 & 0.1566 & 0.1788 \\
    \Xhline{3\arrayrulewidth}
    \rule{0pt}{10pt} \multirow{14}{*}{\makecell[c]{Yes}} & \textbf{Ours}$_\text{syn}$ \cellfirst  &  \textbf{0.0161} &  \textbf{0.0376} &  \textbf{0.0163} &  \textbf{0.0371} &  \textbf{0.0170} &  \textbf{0.0376} \\
     & D.P.  \cellfirst & 0.0251 &  0.0422 &  0.0253 &  0.0422 & 0.0249 & 0.0422 \\
     & BPNet \cellfirst & 0.1494 & 0.2106 & 0.1493 & 0.2107 & 0.1491 &  0.2100 \\
    \rule{0pt}{10pt}  
    & ARKit Depth  \cellfirst & 0.0251 & 0.0424 & 0.0250 & 0.0423  & 0.0254  & 0.0426 \\
    \cline{2-8} 
    \rule{0pt}{10pt} & 
    DepthAny. v2  \cellsecond & 0.0716 & 0.1686 & 0.0616 & 0.1368 & 0.0494 & 0.0764 \\
    & DepthAny. v1  \cellsecond & 0.0733 & 0.1757 & 0.0653 & 0.1530 & 0.0527  & 0.0859 \\
    & Metric3D v2  \cellsecond & 0.0626 & 0.2104 & 0.0524 & 0.1721 & 0.0402  & 0.1045 \\
    & ZoeDepth  \cellsecond & 0.1007 & 0.1917 & 0.0890 & 0.1627 & 0.0762 & 0.1135 \\
    & Lotus  \cellsecond & 0.0624 & 0.0970 & 0.0621 & 0.0962 & 0.0622 & 0.0965  \\
    & Marigold  \cellsecond & 0.0908  & 0.1849 & 0.0807 & 0.1565 & 0.0692 & 0.1065\\
    \cline{2-8} 
    \rule{0pt}{10pt}
      & Metric3D v2  \cellthird & 0.1777 & 0.2766 & 0.1663 & 0.2491 & 0.1615 & 0.2131 \\
      & ZoeDepth  \cellthird & 0.6158 & 0.9577 & 0.5688 & 0.6129 & 0.5316 & 0.5605 \\
    \cline{2-8} 
    \Xhline{3\arrayrulewidth}
    \end{tabular}
  }
  \caption{\textbf{Quantitative comparisons on ARKitScenes dataset.} 
  The terms \textfirst{Net.}, \secondtext{Post.} and \thirdtext{w/o LiDAR} refer to the LiDAR depth usage of models, where ``Net.'' denotes network fusion, ``Post.'' indicates post-alignment using RANSAC, and  ``w/o LiDAR'' means the output is metric depth.
  Methods marked with $^*$ are finetuned with their released models and code on ARKitScenes~\cite{baruch2021arkitscenes} and ScanNet++~\cite{yeshwanth2023scannet++} datasets. 
  }
  \label{tab:arkitscenes}
  \vspace{-8mm}
  \end{center}
\end{table}

\begin{figure*}[t]
    \centering 
    \includegraphics[width=0.92\textwidth]{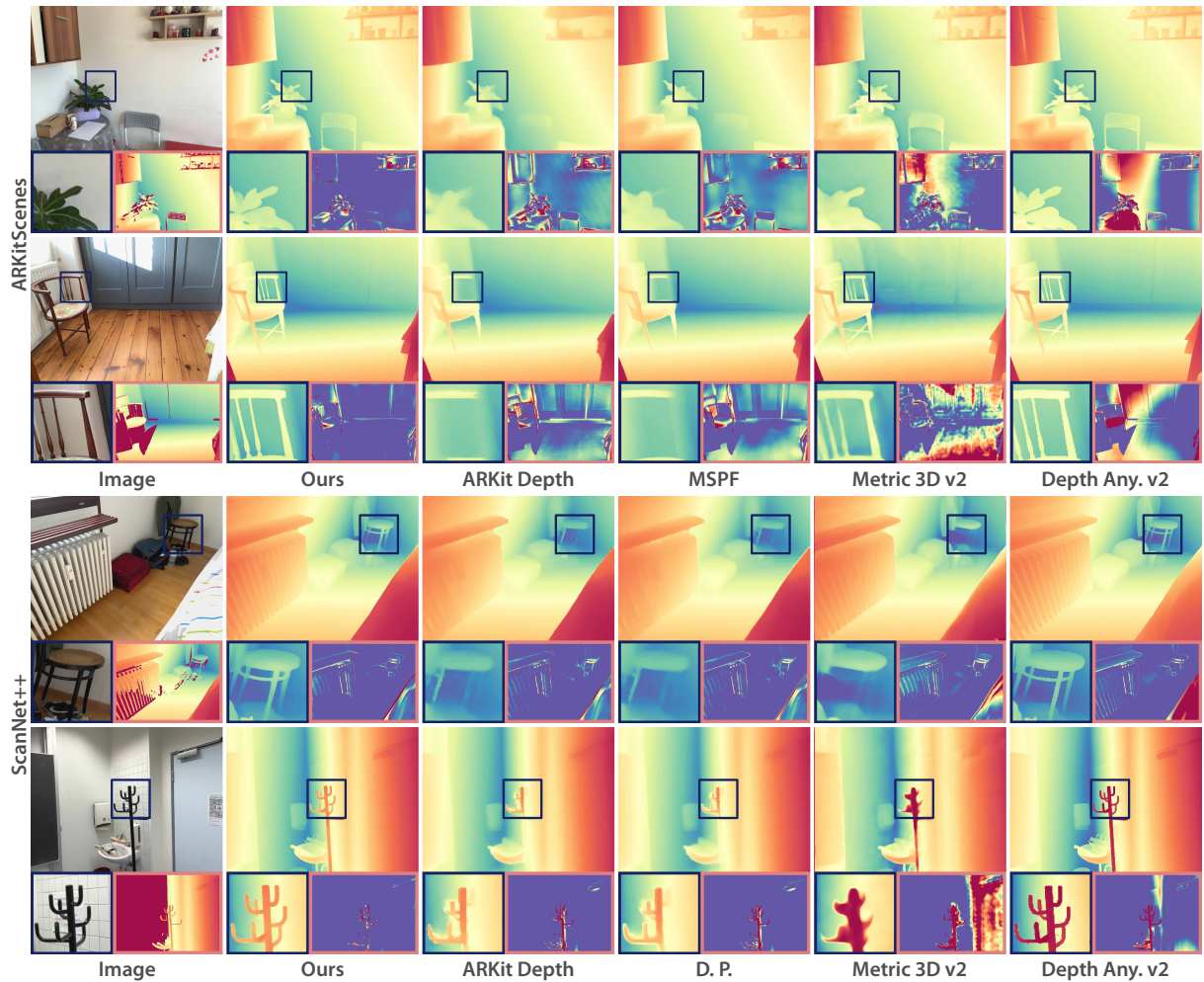}
    \vspace{-6pt}
    \caption{\textbf{Qualitative comparisons with the state-of-the-art.}
    ``Metric3D v2'' and ``Depth Any. v2'' are scale-shift corrected with ARKit depth.
    \textcolor[HTML]{E38182}{The pink boxes} denote the GT depth and depth percentage error map, where \textcolor[HTML]{9B1648}{red} represents high error, and \textcolor[HTML]{6457A8}{blue} indicates low error.
    }
    \label{fig:comp}
\end{figure*}
\begin{figure*}[hbt]
    \centering 
    \includegraphics[width=0.92\textwidth]{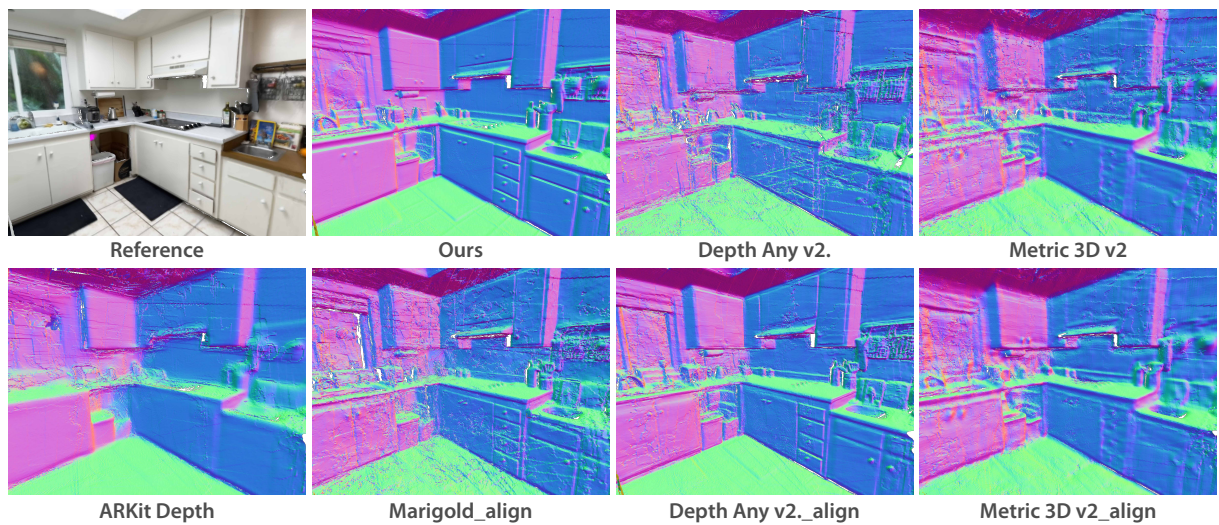}
    \vspace{-6pt}
    \caption{\textbf{Qualitative comparisons of TSDF reconstruction.} *\_align denotes the scale-shift corrected depth with ARKit depth.
    }
    \label{fig:comprecon}
\end{figure*}

\begin{table*}
    \begin{center}
    \resizebox{0.85\linewidth}{!}{
    \begin{tabular}{l|l|cccc|ccccc}
      \Xhline{3\arrayrulewidth}
      \multirow{2}{*}{\makecell[c]{Zero \\ Shot}}  & \multirow{2}{*}{\makecell[c]{\textfirst{Net.} / \secondtext{Post.}/ \\ \thirdtext{w/o LiDAR}}} & \multicolumn{4}{c|}{Depth Estimation}  &  \multicolumn{5}{c}{TSDF Reconstruction} \\
       &   & L1 $\downarrow$ & RMSE $\downarrow$ & AbsRel  $\downarrow$ & $\delta_{0.5}$ $\uparrow$ &  Acc $\downarrow$ & Comp $\downarrow$ & Prec $\uparrow$ & Recall $\uparrow$ & F-score $\uparrow$ \\
      \Xhline{3\arrayrulewidth}
      \rule{0pt}{9pt} \multirow{6}{*}{\makecell[c]{No}} & 
      \cellfirst \textbf{Ours} & \textbf{0.0250} & \textbf{0.0829} & \textbf{0.0175} & \textbf{0.9781} & \textbf{0.0699} & \textbf{0.0616} &  \textbf{0.7255} & \textbf{0.8187} & \textbf{0.7619} \\ 
     & \cellfirst MSPF$^*$ & 0.0326 & 0.0975 & 0.0226 & 0.9674 & 0.0772 & 0.0695 & 0.6738 & 0.7761 & 0.7133 \\
      \cline{2-11}
      & \cellsecond  DepthAny. v2$^*$ &  0.0510 &  0.1010 & 0.0371 & 0.9437 &  0.0808 &0.0735  & 0.6275 &   0.7107 & 0.6595\\
      & \cellsecond ZoeDepth$^*$ &  0.0582 & 0.1069 & 0.0416 & 0.9325 &  0.0881 & 0.0801 & 0.5721 & 0.6640 & 0.6083  \\
      \cline{2-11}
       & \cellthird DepthAny. v2$^*$ & 0.0903 & 0.1347 &  0.0624 & 0.8657 &  0.1264 & 0.0917 & 0.4256 & 0.5954 & 0.4882 \\
       & \cellthird ZoeDepth$^*$  &  0.1675  & 0.1984 &  0.1278 & 0.5807 & 0.1567 & 0.1553 & 0.2164 & 0.2553 & 0.2323 \\
       \Xhline{3\arrayrulewidth}
      \rule{0pt}{9pt} \multirow{9}{*}{\makecell[c]{Yes}}   
      & \cellfirst \textbf{Ours}$_\text{syn}$  & \textbf{0.0327} & \textbf{0.0966} & \textbf{0.0224} & \textbf{0.9700}  & \textbf{0.0746} & \textbf{0.0666} & \textbf{0.6903} & \textbf{0.7931} & \textbf{0.7307} \\ 
       & \cellfirst D.P. & 0.0353  & 0.0983 & 0.0242 & 0.9657 & 0.0820 & 0.0747 & 0.6431 & 0.7234 & 0.6734 \\
       & \cellfirst ARKit Depth & 0.0351 & 0.0987 & 0.0241 &  0.9659 & 0.0811 & 0.0743 & 0.6484  & 0.7280  &  0.6785 \\
       \cline{2-11}
    & \cellsecond DepthAny. v2 &  0.0592  & 0.1145 & 0.0402  &   0.9404  & 0.0881 & 0.0747 & 0.5562 & 0.6946 & 0.6127 \\ 
    & \cellsecond Depth Pro &  0.0638 & 0.1212 & 0.0510  & 0.9212  & 0.0904 & 0.0760 & 0.5695 & 0.6916 & 0.6187 \\ 
    & \cellsecond Metric3D v2 & 0.0585  &  0.3087  & 0.0419 & 0.9529 & 0.0785 & 0.0752 & 0.6216 & 0.6994 & 0.6515   \\ 
    & \cellsecond Marigold & 0.0828 &  0.1412 & 0.0603 & 0.8718 & 0.0999 & 0.0781 & 0.5128 & 0.6694 & 0.5740  \\ 
    \cline{2-11}
     & \cellthird DepthPro  & 0.2406 & 0.2836 & 0.2015 &  0.5216 &  0.1537 & 0.1467 & 0.2684 & 0.3752 & 0.3086 \\ 
     & \cellthird Metric3D v2 & 0.1226 & 0.3403 &  0.0841 & 0.8009   &  0.0881 & 0.0801 & 0.5721 & 0.6640 & 0.6083 \\ 
      \Xhline{3\arrayrulewidth}
      \end{tabular}
    }
    \vspace{-2mm}
    \caption{\textbf{Quantitative comparisons on ScanNet++ dataset.}
    The terms \textfirst{Net.}, \secondtext{Post.} and \thirdtext{w/o LiDAR} refer to the LiDAR depth usage of models as the last table.
  Methods marked with $^*$ are finetuned with their released  code on ARKitScenes~\cite{baruch2021arkitscenes} and ScanNet++~\cite{yeshwanth2023scannet++} datasets. 
    }
    \label{tab:scannetpp}
    \vspace{-6mm}
    \end{center}
\end{table*}
\begin{table}
    \setlength\tabcolsep{2 pt}
    \begin{center}
    \resizebox{\columnwidth}{!}{
    \begin{tabular}{l|cc|ccc}
      \Xhline{3\arrayrulewidth}
       &   \multicolumn{2}{c|}{ARKitScenes}  &  \multicolumn{3}{c}{ScanNet++} \\
      & L1 $\downarrow$ & AbsRel $\downarrow$ & Acc $\downarrow$ & Comp $\downarrow$  & F-Score $\uparrow$ \\
       \Xhline{3\arrayrulewidth}
       (a) Ours$_\text{syn}$ (synthetic data) &  0.0163 &  0.0142 &  0.0746 &  0.0666 &  0.7307 \\
       (b) w/o prompting                      &  0.0605 &  0.0505 &  0.0923 &  0.0801 &  0.5696 \\
       (c) w/o foundation model               &  0.0194 &  0.0169 &  0.0774 &  0.0713 &  0.7077 \\
       \hline
       (d) AdaLN prompting                    &  0.0197 &  0.0165 &  0.0795 &  0.0725 &  0.6943 \\
       (e) Cross-atten. prompting             &  0.0523 &  0.0443 &  0.0932 &  0.0819 &  0.5595 \\
       (f) Controlnet prompting               &  0.0239 &  0.0206 &  0.0785 &  0.0726 &  0.6899 \\
       \hline
       (g) a + ARKitScenes data                   &  0.0134 &  0.0115 &  0.0744 &  0.0662 &  0.7341 \\
       (h) g + ScanNet++ anno. GT                 &  0.0132 &  0.0114 &  0.0670 &  0.0614 &  0.7647 \\
       (i) g + ScanNet++ pseudo GT                &  0.0139 &  0.0121 &  0.0835 &  0.0766 &  0.6505 \\
       (j) \textbf{Ours} (h,i+edge loss)           &  0.0132 &  0.0115 &  0.0699 &  0.0616 &  0.7619 \\
      \Xhline{3\arrayrulewidth}
      \end{tabular}
    }
    \vspace{-2mm}
    \caption{\textbf{Quantitative ablations on ARKitScenes and ScanNet++ datasets.} Please refer to \cref{sec:ablation} for detailed descriptions.
    }
    \label{tab:ablation}
    \vspace{-7mm}
    \end{center}
  \end{table}

\section{Experiments}
\subsection{Experimental Setup}

We mainly conduct experiments on the HyperSim synthetic dataset~\cite{roberts2021} and two real-world datasets: ScanNet++~\cite{yeshwanth2023scannet++} and ARKitScenes~\cite{baruch2021arkitscenes}, which provide iPhone RGB-LiDAR data~($192 \times 256$ resolution) and annotated depth from a high-power LiDAR~($1440 \times 1920$ resolution). 
We follow the suggested training and evaluation protocol in~\cite{baruch2021arkitscenes} for ARKitScenes, where 40K images are used for training and 5K images for evaluation.
For the ScanNet++ dataset, we randomly select 20 scenes from its 50 validation scenes, amounting to approximately 5K images for our validation and the training set are from its 230 training scenes, containing about 60K images.
To ensure a fair comparison, we additionally train a model with HyperSim training set to achieve zero-shot testing on ScanNet++ and ARKitScenes datasets.
Besides depth accuracy metrics, we also report the TSDF reconstruction results of our method on ScanNet++, which reflects the depth consistency. We describe the details of the evaluation metrics in the supp.



\subsection{Comparisons with the State of the Art}
We compare our method against the current SOTA depth estimation methods from two classes: \textit{Monocular depth estimation (MDE)} and \textit{depth completion/upsampling}. For MDE methods, we compare our method with Metric3D v2~\cite{hu2024metric3dv2}, ZoeDepth~\cite{bhat2023zoedepth}, DepthPro~\cite{bochkovskii2024depth}, Depth Anything v1 and v2~\cite{yang2024depth,yang2024depthv2} (short for DepthAny. v1 and v2), Marigold~\cite{ke2024repurposing} and Lotus~\cite{he2024lotus}. 
For depth completion/upsampling methods, we compare our method with BPNet~\cite{BP-Net}, Depth Prompting~\cite{park2024depth} (short for D.P.), MSPF~\cite{xian2020multi}. 
To make a fair comparison with MDE methods, we align their predictions with ARKit LiDAR depth using the RANSAC align method.
According to whether they have seen the testing data types during training, we divide methods into two categories: \textit{zero-shot} and \textit{non zero-shot}.
We train a model Ours$_\text{syn}$ only with HyperSim training set to make comparisons with the zero-shot methods.
As shown in \cref{tab:arkitscenes,tab:scannetpp,fig:comp,fig:comprecon}, our method consistently outperforms the existing methods. Note that Ours$_\text{syn}$ achieves better performance than all non-zero-shot models~\cite{yang2024depthv2,xian2020multi} on ScanNet++, highlighting the generalization ability of prompting a depth foundation model.

\subsection{Ablations and Analysis}
\label{sec:ablation}

\paragraph{Prompting a depth foundation model.}
We assess its importance with two experiments:
1) Removing the prompting. \cref{tab:ablation}\citecolor{(b)} shows a significant performance drop.
2) Removing the foundation model initialization~\cite{yang2024depthv2}. \cref{tab:ablation}\citecolor{(c)} shows a noticeable performance decline.

\paragraph{Prompting architecture design.} We study different designs: AdaLN, Cross-attention, and ControlNet as discussed in \cref{sec:method:fusion}. \cref{tab:ablation}\citecolor{(d,e,f)} reveals that ControlNet performs best but still falls short of our method.


\paragraph{Training data and edge-aware depth loss.}
We initially incorporate ARKitScenes data, which only enhances performance on ARKitScenes (\cref{tab:ablation}\citecolor{(g)}). 
Then we add ScanNet++, which improves results on both ARKitScenes and ScanNet++ (\cref{tab:ablation}\citecolor{(h)}). However, the depth visualization remains less than ideal (\cref{fig:ablation}). 
\cref{tab:ablation}\citecolor{(i)} show that direct supervision with pseudo GT depth from reconstruction methods decreases performance.
Ultimately, employing the edge-aware depth loss that utilizes pseudo GT depth and FARO annotated GT achieves comparable performance with ~\cref{tab:ablation}\citecolor{(h)} but with superior thin structure depth performance as shown in \cref{fig:ablation}. We provide more qualitative ablation results in the supp.

\begin{figure}[t]
    \centering 
    \includegraphics[width=1\columnwidth]{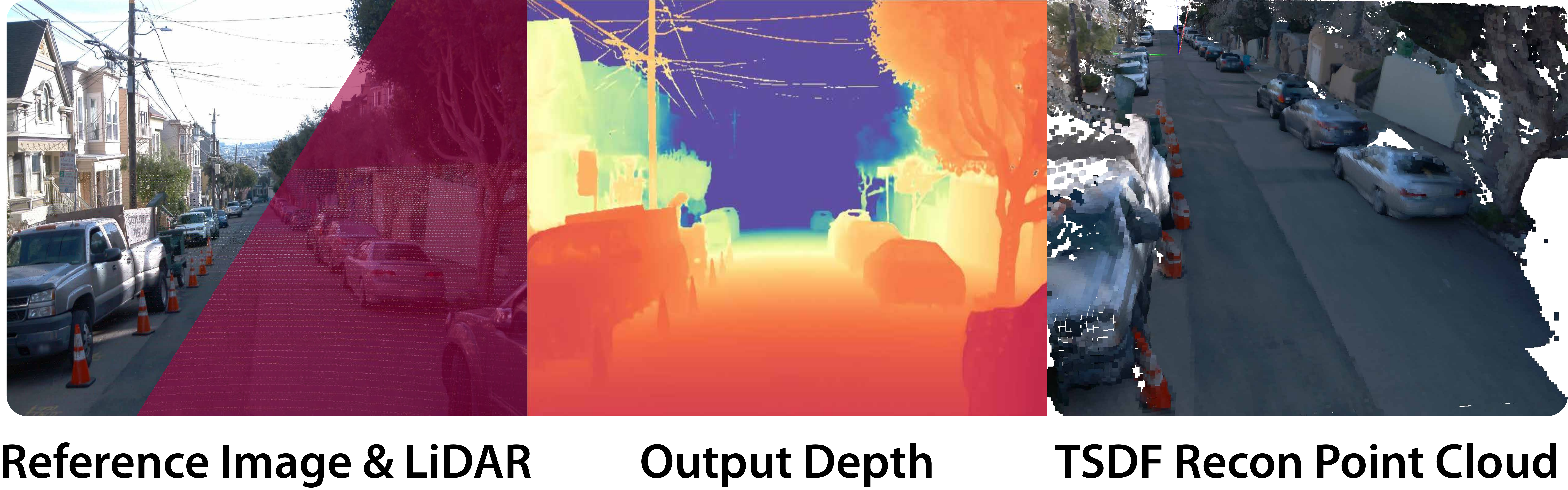}
    \caption{\textbf{Outdoor reconstruction by taking the vehicle LiDAR as metric prompt.} Please refer to the supp. for more video results.
    }
    \label{fig:driving}
    \vspace{-8pt}
\end{figure}



\paragraph{Running time analysis.} Our model with ViT-L runs at 20.4 FPS for an image resolution of $768 \times 1024$ on a A100 GPU. As ARKit6 supports 4K image recording, we test our model at a resolution of $2160 \times 3840$ and achieve 2.0 FPS. Note that our model can also be implemented with ViT-S, where the corresponding speeds are 80.0 and 10.3 FPS. More testing results can be found in the supp.

\begin{figure*}[t]
    \centering 
    \includegraphics[width=0.99\textwidth]{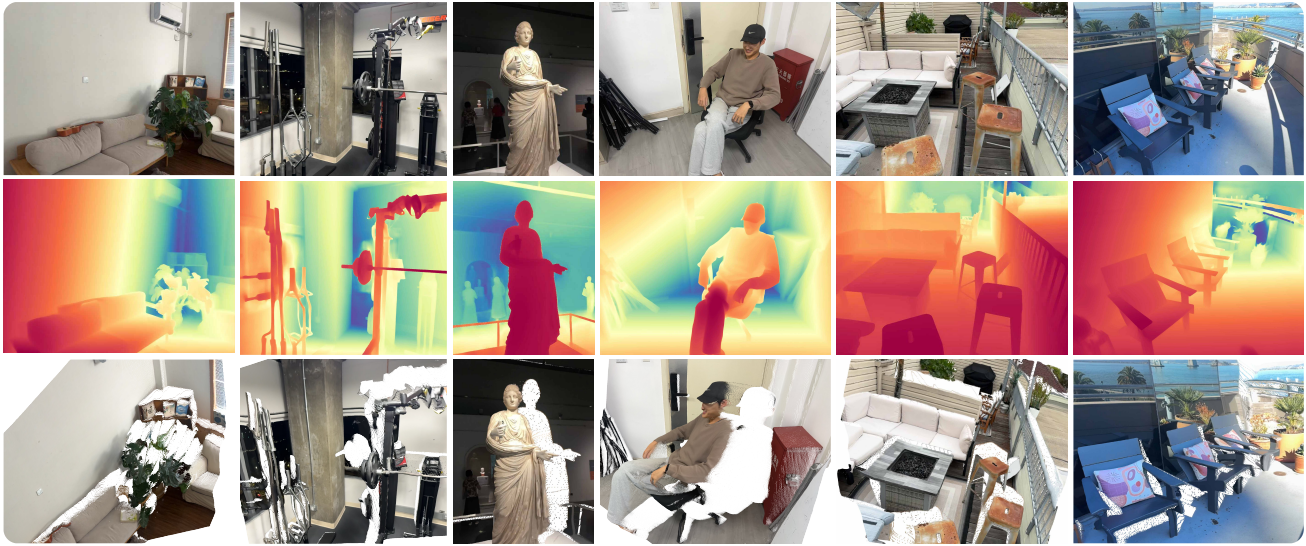}
    \vspace{-2mm}
    \caption{\textbf{Zero-shot testing on diverse scenes.}}
    \vspace{-4mm}
    \label{fig:zeroshot}
\end{figure*}

\subsection{Zero-shot Testing on Diverse Scenes}
Although our model is trained on indoor scenes, it generalizes well to various scenarios, including new rooms, gyms with thin structures, poorly lit museums, human and outdoor environments, as shown in \cref{fig:zeroshot}, highlighting the effectiveness of prompting a depth foundation model. 
Please refer to the supp. for video results.


\subsection{Application: 3D Reconstruction}
Our consistent and scale-accurate depth estimation benefits the indoor 3D reconstruction as shown in \cref{tab:scannetpp,fig:comprecon}. Besides, the prompt of our model can be easily replaced with vehicle LiDAR, which enables our model to achieve large-scale outdoor scene reconstruction as shown in \cref{fig:driving}. We detail the setup and include more video results for dynamic streets in the supp.

\subsection{Application: Generalized Robotic Grasping}
\begin{figure}[t]
    \centering
    \includegraphics[width=\columnwidth]{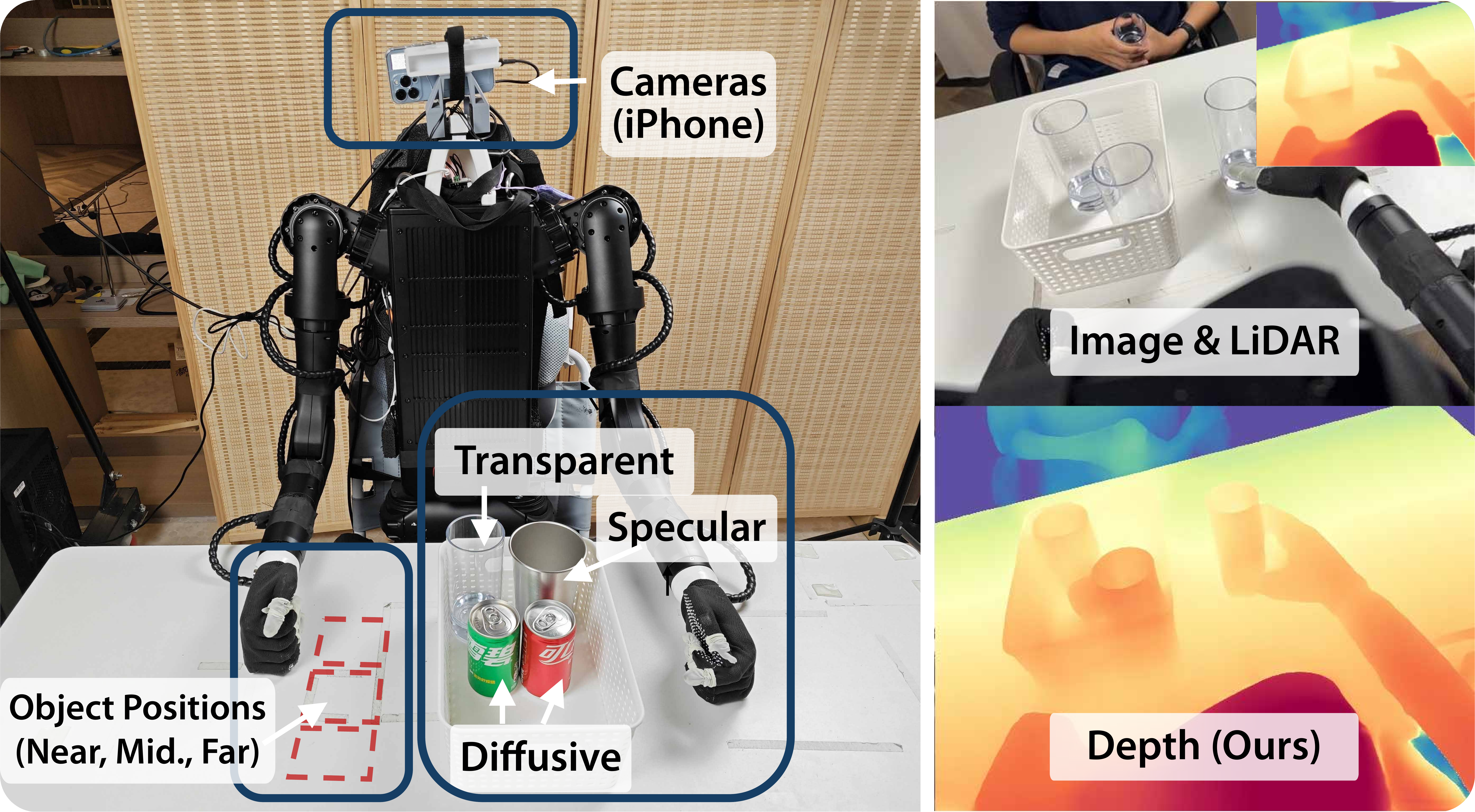}
    \caption{\textbf{Robotic grasping setup and input signal types.} Our goal is to grasp objects of various types using image/LiDAR/depth inputs. Red rectangles indicate potential object positions.
    }
    \label{fig:robot-setup}
    \vspace{-8pt}
\end{figure}
We set up a robotic platform to test our model in generalized robotic manipulation (Fig.~\ref{fig:robot-setup}), which typically requires depth or RGB as observations. Good depth estimation enhances the generalization ability because it accurately describes the 3D information of surroundings~\cite{ze2024generalizable,hua2024gensim2}. Specifically, we train an ACT policy~\cite{zhao2023learning} to grasp various objects into the box, using different types of input signals such as RGB, LiDAR, and depth data from our model.
We empirically find that our model generalizes well to unseen objects like transparent and specular objects when trained on diffusive objects, outperforming RGB and LiDAR inputs as shown in \cref{tab:robot}. This is because RGB is dominated by color, which leads to poor generalization across objects, and the iPhone LiDAR depth is noisy and lacks the capability to perceive transparent objects. Please refer to the supp. for detailed setup descriptions and videos.

\begin{table}[t]
    \centering
    \setlength\tabcolsep{4 pt}
    \resizebox{\columnwidth}{!}{
    \begin{tabular}{l|cc|c|c}
    \Xhline{3\arrayrulewidth}
     \multirow{2}{*}{\makecell[c]{Input Signal}} & \multicolumn{2}{c|}{\textcolor{blue}{Diffusive}} & \multirow{2}{*}{\makecell[c]{Transparent }} & \multirow{2}{*}{\makecell[c]{Specular}} \\
     & Red Can & Green Can & & \\
    \hline
     \textbf{Ours} & \textbf{1.0/1.0/1.0} & \textbf{1.0/1.0/1.0} & \textbf{0.3/1.0/1.0} & \textbf{0.8/1.0/0.9} \\
     LiDAR & \textbf{1.0/1.0/1.0} & 1.0/1.0/0.2 & 0.5/0.4/0.0 & 0.7/1.0/0.0 \\
     RGB & 1.0/1.0/0.0 & 1.0/1.0/0.0 & 0.2/1.0/0.0 & 0.0/0.9/0.9 \\
    \Xhline{3\arrayrulewidth}
    \end{tabular}}
    \vspace{-6pt}
    \caption{\textbf{Grasping success rate on various objects}. Three numbers indicate objects placed at near, middle, and far positions. The grasping policy is trained on \textcolor{blue}{diffusive} and tested on all objects.}
    \label{tab:robot}
    \vspace{-18pt}
\end{table}


\section{Conclusion and Discussions}
This paper introduced a new paradigm for metric depth estimation, formulated as prompting a depth foundation model with metric information. 
We validated the feasibility of the paradigm by choosing the low-cost LiDAR depth as the prompt.
A scalable data pipeline was proposed to generate synthetic LiDAR depth and pseudo GT depth for training.
Extensive experiments demonstrate the superiority of our method against existing monocular depth estimation and depth completion/upsampling methods.
Furthermore, we showed that it benefits for downstream tasks including 3D reconstruction and generalized robotic grasping.

\paragraph{Limitations and future work.} This work has some known limitations. 
For instance, when using the iPhone LiDAR as the prompt, it cannot handle long-range depth, as the iPhone LiDAR detects very noisy depth for far objects.
Additionally, we observed some temporal flickering of LiDAR depth, leading to a flickering depth prediction. 
These issues can be addressed in future works by considering more advanced prompt learning techniques that can extend the effective range and temporal prompt tuning.

{
    \small
    \bibliographystyle{ieeenat_fullname}
    \bibliography{main}
}

\begin{figure*}[t]
\begin{center}
    \includegraphics[width=\linewidth]{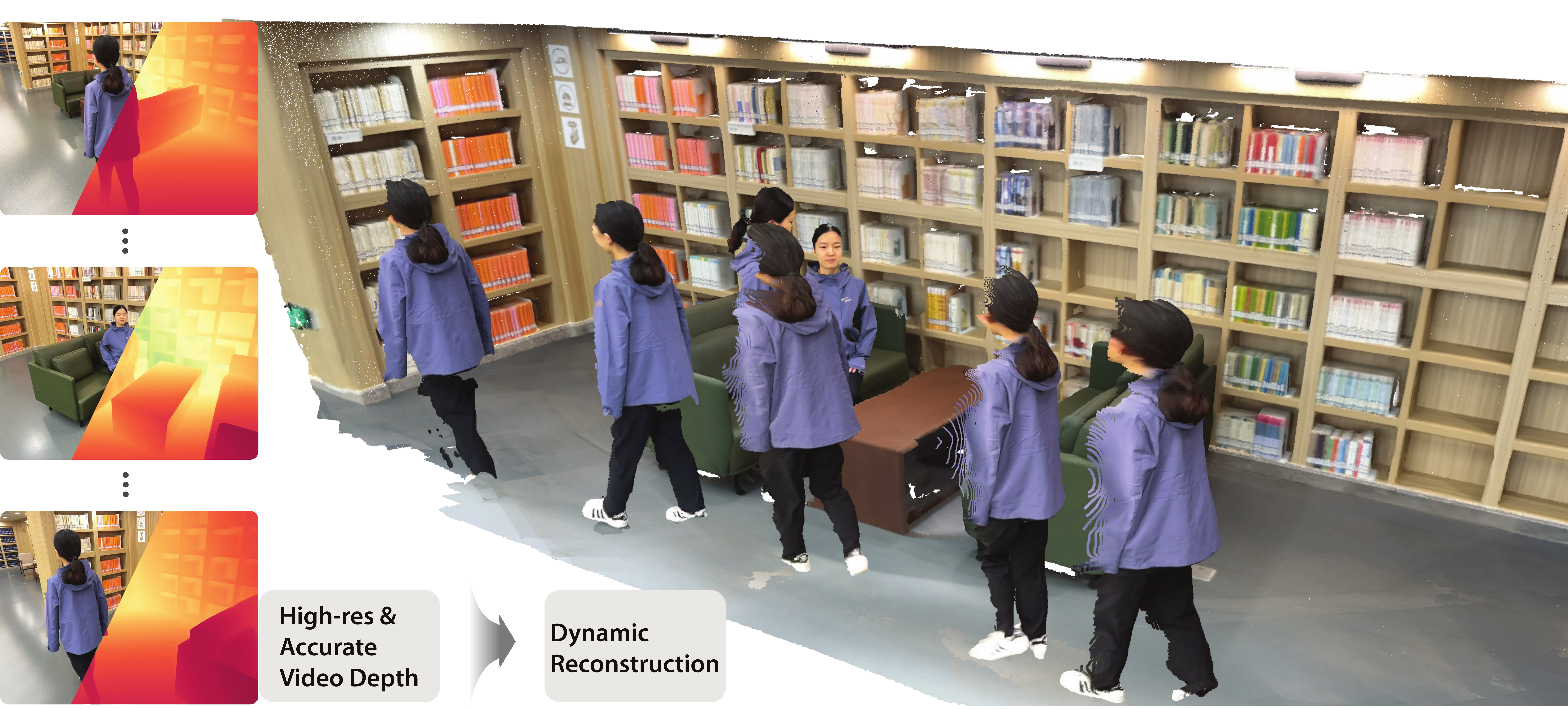}
\end{center}
\caption{
Our accurate and high-resolution depth enables dynamic 3D reconstruction from a single moving camera.
Here we illustrate the reconstruction results of a human walking in the library. The foreground is segmented with a SAM2~\cite{ravi2024sam} model.
}
\label{fig:suppteaser}
\end{figure*}

\newpage

\appendix
In the supplementary material, we present more discussions, additional results, and implementation details. Please find more video results in our supplementary video.

\section{Additional Discussions}

\begin{figure*}[t]
    \centering 
    \includegraphics[width=0.99\textwidth]{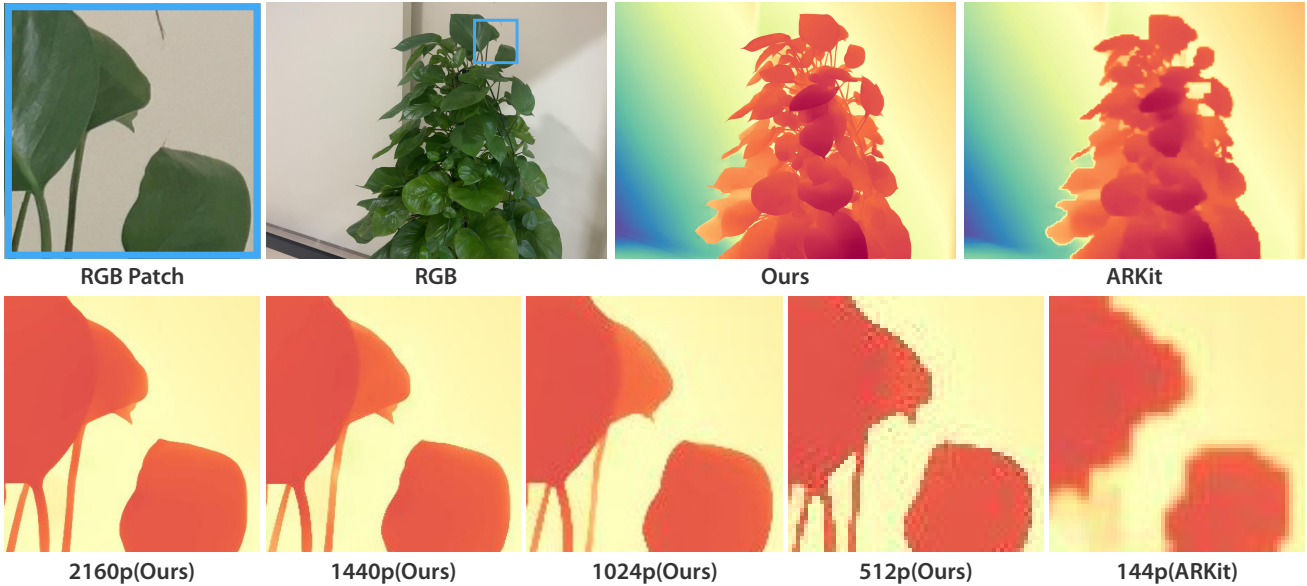}
    \caption{\textbf{Generalizability to different resolutions.}
    Our model can infer depth for images of different resolutions from 512p to 2160p.
    }
    \label{fig:multires}
\end{figure*}

\subsection{Generalizability to Different Resolutions}
This section discusses the generalization capability of our model across different image and lidar depth resolutions provided by ARKit4 and ARKit6. ARKit4 captures images at a maximum resolution of $1440 \times 1920$ at 60Hz and lidar depth at $192 \times 256$, while ARKit6 captures images at a maximum resolution of $3024 \times 4032$ at 30Hz and lidar depth at $240 \times 320$. Both ScanNet++~\cite{yeshwanth2023scannet++} and ARKitScenes~\cite{baruch2021arkitscenes} are collected using ARKit4. Although our model is trained using ScanNet++ and ARKitScenes data at a resolution of $1440 \times 1920$, we find that it generalizes well to ARKit6 images and depth at different resolutions. As shown in \cref{fig:multires}, we include a comparison of depth estimation for images of different resolutions, with an image resolution of $2160 \times 3840$ and a lidar depth resolution of $144 \times 256$, captured from the ARKit6 API.

\subsection{Why Do We Need Synthetic Data?}
The advantages of synthetic data include high-quality ground truth depth, which has been crucial for the success of many recent depth estimation works~\cite{yang2024depthv2,ke2024repurposing,fu2025geowizard,bochkovskii2024depth}. We also utilize synthetic data to achieve high-quality depth estimation results. Furthermore, the availability of real data with lost-cost LiDAR and high-power LiDAR is currently limited~\cite{yeshwanth2023scannet++,baruch2021arkitscenes}, primarily to indoor scenes, while synthetic data can further enhance diversity; for instance, our experiments have shown that including human synthetic data~\cite{tao2021function4d} improves our method's generalization to human subjects.

\begin{figure}[htb]
    \centering 
    \includegraphics[width=0.99\columnwidth]{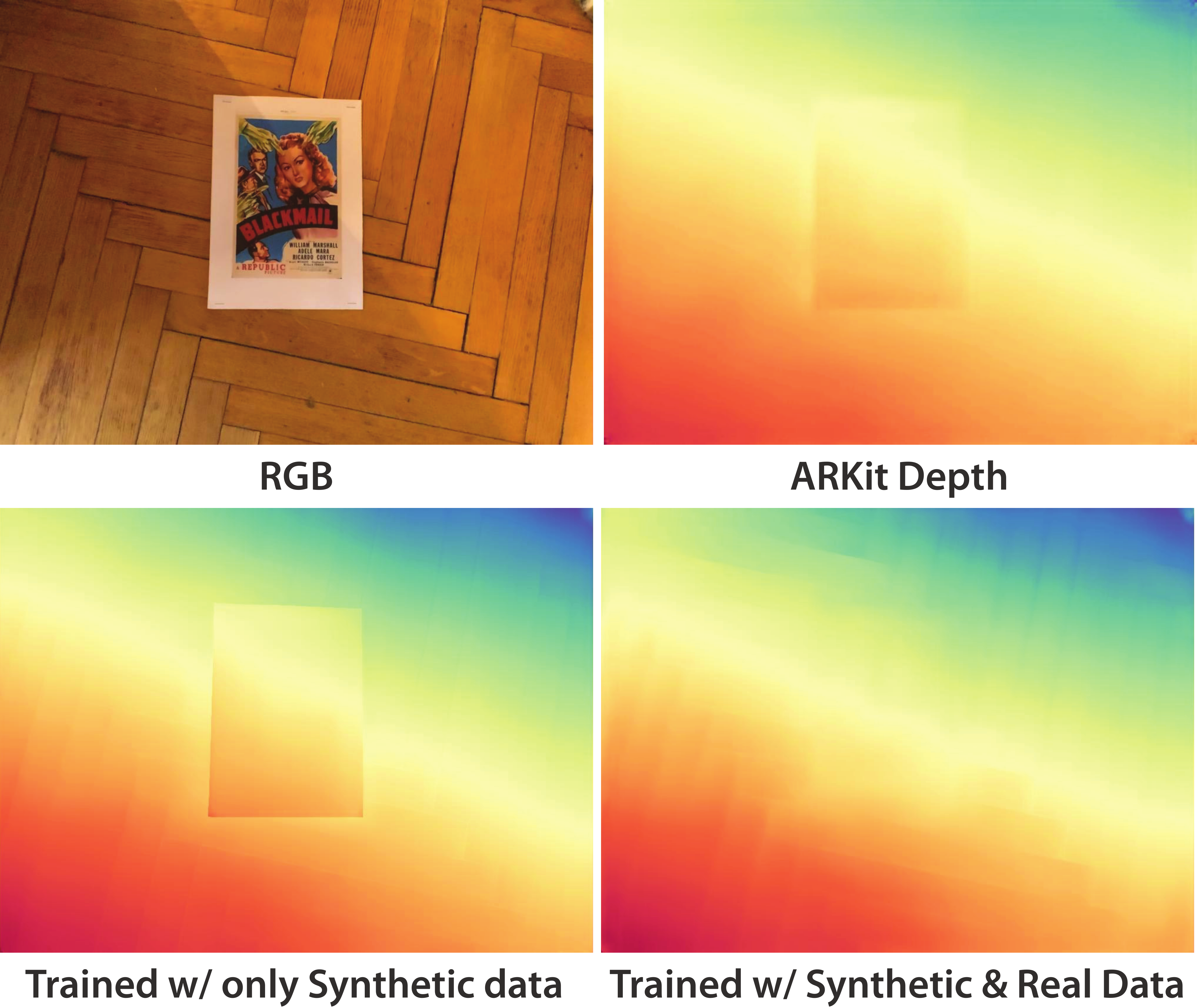}
    \caption{\textbf{Effects of using real data.} 
    }
    \label{fig:realdata}
\end{figure}

\subsection{Why Do We Need Real Data?}
Training with real data can further address the inability of synthetic LiDAR simulation to replicate LiDAR noise patterns, thereby enhancing depth estimation capabilities.
By utilizing synthetic data, we have achieved preliminary results. However, as demonstrated by the quantitative experiments in the main paper, the use of real data further enhances the performance. Here, we include additional qualitative results in \cref{fig:realdata}, which show that real data is beneficial because LiDAR simulation methods cannot fully replicate the noise of real LiDAR.

\begin{table}
    \setlength\tabcolsep{2 pt}
    \begin{center}
    \resizebox{\columnwidth}{!}{
    \begin{tabular}{l|cc|ccc}
      \Xhline{3\arrayrulewidth}
       &   \multicolumn{2}{c|}{ARKitScenes}  &  \multicolumn{3}{c}{ScanNet++} \\
      & L1 $\downarrow$ & AbsRel $\downarrow$ & Acc $\downarrow$ & Comp $\downarrow$  & F-Score $\uparrow$ \\
       \Xhline{3\arrayrulewidth}
       (a) Depth Any. as foundation &  0.0132 &  0.0115 &  0.0699 &  0.0616 &  0.7619 \\
       (b) Depth Pro as foundation &  0.0169 &  0.0150 &  0.0754 &  0.0676 &  0.7202 \\
       (c) Depth Pro &  0.1225 &  0.1038 &  0.0904 & 0.0760 &  0.6187 \\
      \Xhline{3\arrayrulewidth}
      \end{tabular}
    }
    \caption{\textbf{Additional quantitative ablations.} 
    Please refer to \cref{sec:depthfoud} for detailed descriptions.}
    \label{tab:suppablation}
    \end{center}
  \end{table}
\subsection{Replacing Depth Foundation Models}
\label{sec:depthfoud}
Since our model is a general design for DPT, it can be easily adapted to other depth foundation models that also utilize the DPT structure, such as Depth Pro~\cite{bochkovskii2024depth}.
Our experiments demonstrate that it significantly enhances the performance of Depth Pro, as shown in \cref{tab:suppablation}\citecolor{-(b,c)}, although it does not outperform our choice of Depth Anything~\cref{tab:suppablation}\citecolor{-(a)}. 

\section{Additional Results}

\begin{figure}[htb]
    \centering 
    \includegraphics[width=0.99\columnwidth]{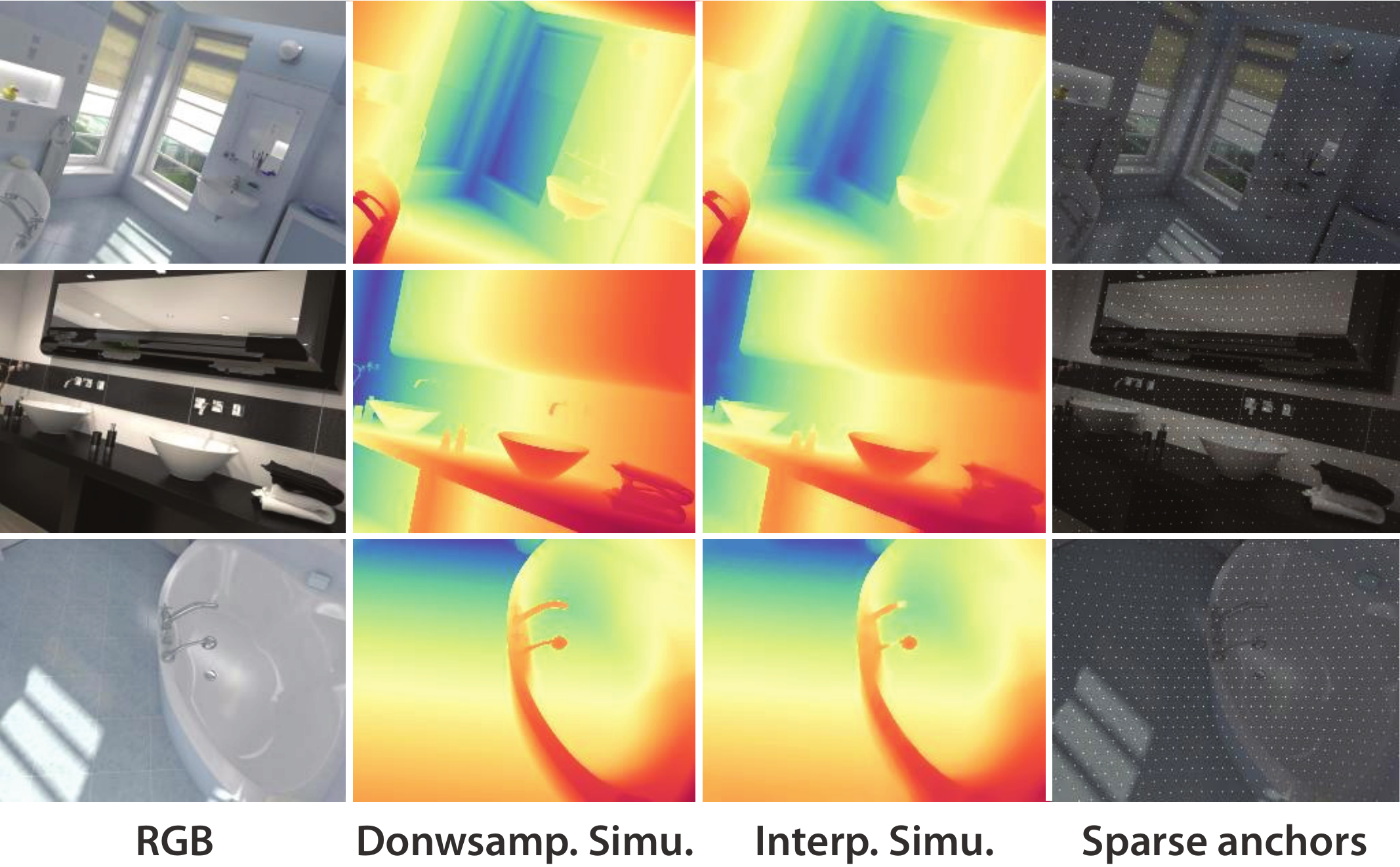}
    \caption{\textbf{Visualization results of simulated LiDAR.} 
    ``Interp. Simu.'' is the proposed interpolation method, which is interpolated from sparse anchors depth. This method effectively simulates the noise of real LiDAR data. We also provide the naive downsampled simulated LiDAR for comparison.
    }
    \label{fig:lidarsimulation}
\end{figure}

\paragraph{Visualization results of simulated LiDAR.}
We provide the visualization results of our simulated LiDAR in \cref{fig:lidarsimulation}.

\begin{figure}[htb]
    \centering 
    \includegraphics[width=0.99\columnwidth]{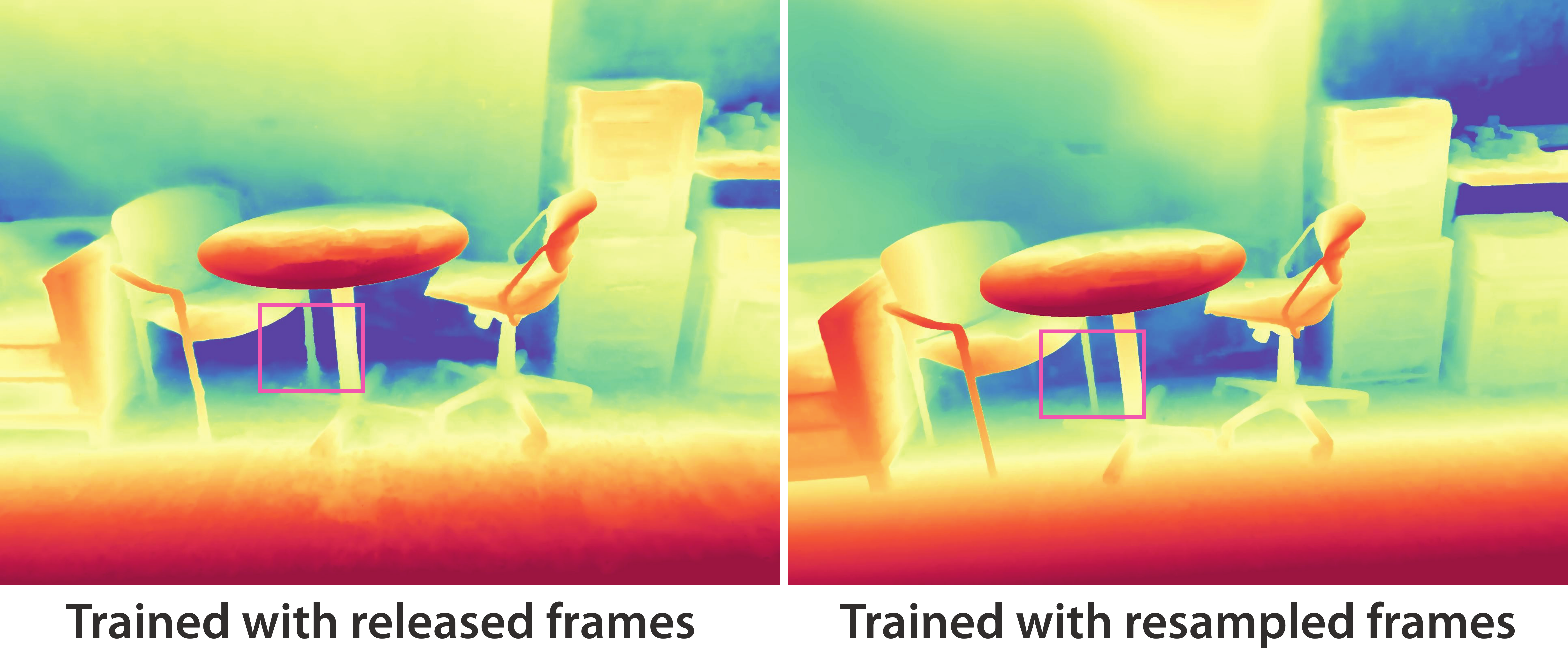}
    \caption{\textbf{ZipNeRF depth of different training frames.}
    Training with resampled frames removing blurred frames leads to a better ZipNeRF reconstruction. 
    }
    \label{fig:unblurframes}
\end{figure}

\paragraph{ZipNeRF reconstruction results.}
High-quality and dense observations are essential for effective 3D reconstruction.
However, the iPhone data from ScanNet++~\cite{yeshwanth2023scannet++} frequently exhibits motion blur. To address this, we resample videos from ScanNet++ to remove blurring frames.
Specifically, we calculate the variance of Laplacians for each image to assess its sharpness and use the sharpness score to select frames. For a 60fps video, we select one frame every 30 frames, ensuring no repeated selection within any 6 consecutive frames, and guarantee at least one selection within every 2 seconds.
We find that this method significantly reduces motion blur and leads to a better ZipNeRF reconstruction as shown in \cref{fig:unblurframes}.
Additionally, we utilize both the DSLR and iPhone data released by ScanNet++ to optimize ZipNeRF, which substantially improved our experimental results.
Training ZipNeRF on ScanNet++ data required approximately $280 \times 2.5 \times 8$ GPU hours. We will release our processed data to benefit the research community.

\begin{figure}[htb]
    \centering 
    \includegraphics[width=0.99\columnwidth]{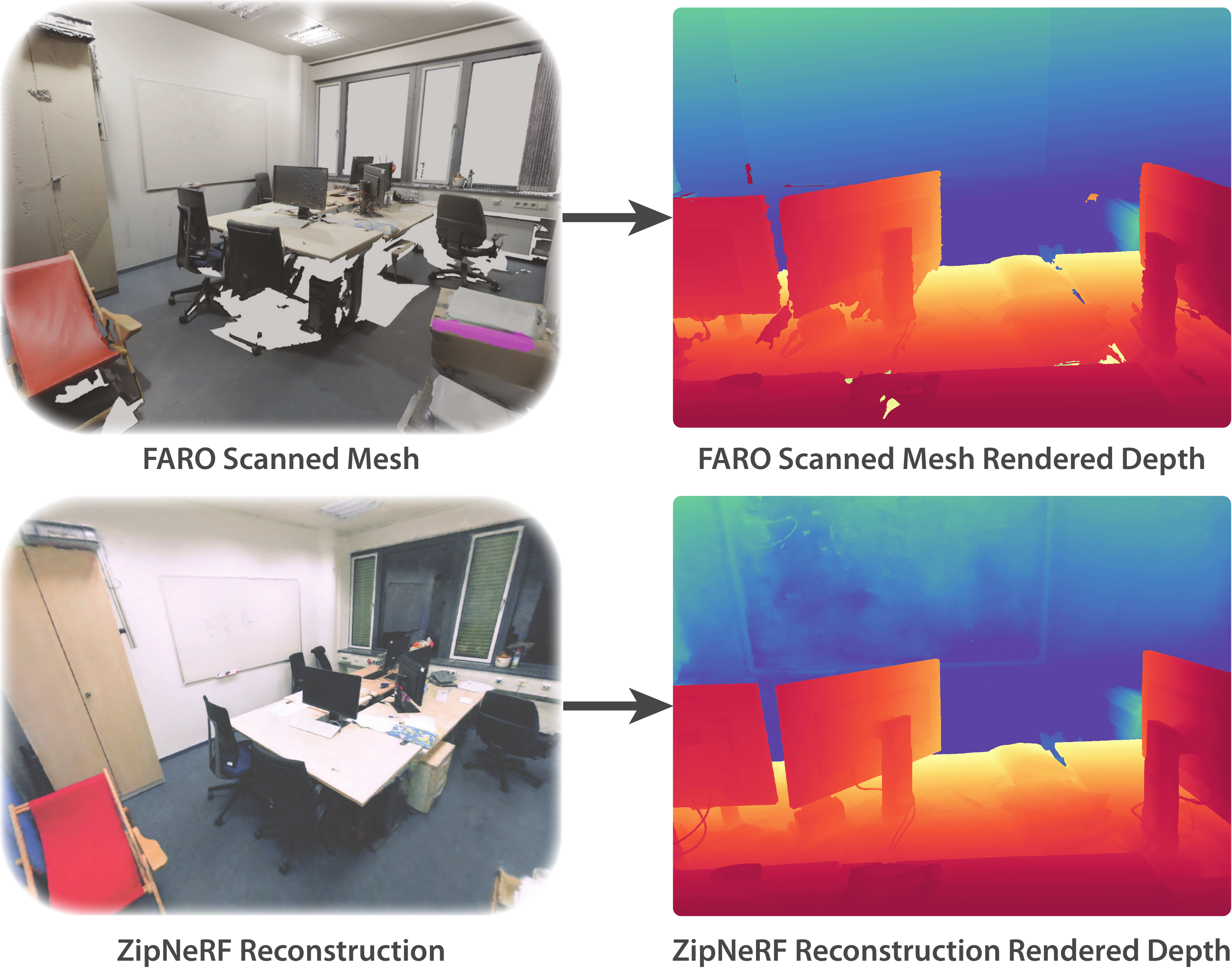}
    \caption{\textbf{Illustration of different depth annotation types.} 
    Please refer to \cref{sec:meshnerf} for more descriptions.
    }
    \label{fig:meshnerf}
\end{figure}

\paragraph{Illustration of different annotation types.}
\label{sec:meshnerf}
We provide an illustration of different annotation types in \cref{fig:meshnerf}.
Here we clearly observe the issues and advantages of different depth annotation types.
The GT depth in ScanNet++ is annotated using a FARO scanned mesh.
Due to the presence of many occlusions in the scene, the scanned mesh is incomplete, resulting in depth maps with numerous holes and poor edge quality.
The pseudo GT depth annotated using NeRF reconstruction has accurate edges but performs poorly in planar regions.
Therefore, an edge-aware loss is proposed to merge their advantages.

\section{More Details}

\subsection{Details about Our Model}
We employ the ViT-large model from Depth Anything v2~\cite{yang2024depthv2} as our backbone model. 
The shallow convolutional network consists of two convolutional layers, each with a kernel size of 3 and a stride of 1, utilizing ReLU as the non-linear activation function.
The zero-initialized projection layer is a $1\times1$ convolutional layer.
For training on the ScanNet++~\cite{yeshwanth2023scannet++} dataset, we apply the loss function proposed in the main paper.
For training on the ARKitScenes~\cite{baruch2021arkitscenes} dataset, we exclusively use the L1 loss.
For training on synthetic~\cite{roberts2021} data, we employ both gradient and pixel-wise L1 loss simply from ground-truth depth supervision.

\begin{figure*}[t]
    \centering 
    \includegraphics[width=0.99\textwidth]{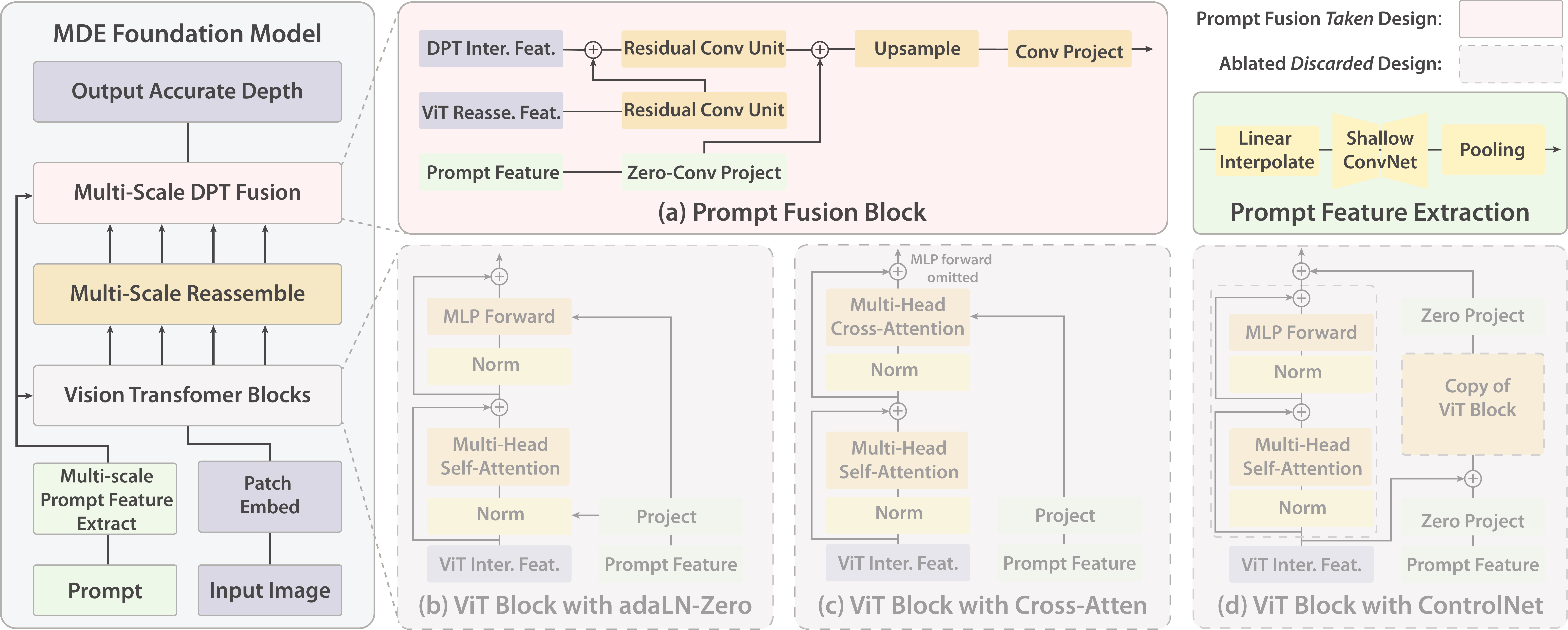}
    \caption{\textbf{Illustrations of our method and optional designs.}
     Please refer to \cref{sec:optional} for more details.
    }
    \label{fig:supppipe}
\end{figure*}

\subsection{Optional Design Details}
\label{sec:optional}
As mentioned in the main paper, in addition to the proposed design, we also explore optional designs including AdaLN~\cite{perez2018film}, Cross-attention~\cite{vaswani2017attention}, and ControlNet~\cite{zhang2023adding}.
We include a figure to illustrate these designs in \cref{fig:supppipe}.
Our experiments (Tab.~\citecolor{3} in the main paper) show that ControlNet performs the best among these alternatives, but it is still not as effective as our proposed design.
The plausible reason is that they are designed to integrate cross-modal information (e.g., text prompts), which does not effectively utilize the pixel alignment characteristics between the input low-resolution depth and the output depth. 
We also combine the proposed design with ControlNet to investigate potential further improvements.
However, no additional improvements are observed~(ours vs. combination are 0.730 vs. 0.731 in terms of F-score metric on ScanNet++), but the computational costs increase. Therefore, we keep the proposed design in the final version.

\subsection{Evaluation Metrics}
For depth metrics, we report L1, RMSE, AbsRel and $\delta_{0.5}$. Their definitions can be found in \cref{tab:depthmetric}.
\begin{table}[ht]
    \centering
    \begin{tabular}{lc}
    \toprule
    Metric & Definition \\
    \midrule
    \vspace{3mm}
    L1 & $\frac{1}{N} \sum_{i=1}^{N} |\mathbf{D}_i - \hat{\mathbf{D}}_i|$ \\
    \vspace{3mm}
    RMSE & $\sqrt{\frac{1}{N} \sum_{i=1}^{N} (\mathbf{D}_i - \hat{\mathbf{D}}_i)^2}$ \\
    AbsRel & $\frac{1}{N} \sum_{i=1}^{N} |\mathbf{D}_i - \hat{\mathbf{D}}_i|/\mathbf{D}_i$ \\
    $\delta_{0.5}$ & $\frac{1}{N} \sum_{i=1}^{N} \mathbb{I} \left( \max \left( \frac{\mathbf{D}_i}{\hat{\mathbf{D}}_i}, \frac{\hat{\mathbf{D}}_i}{\mathbf{D}_i} \right) < 1.25^{0.5} \right)$ \\
    \bottomrule
    \end{tabular}
    \caption{\textbf{Depth metric definitions.} $\mathbf{D}$ and $\hat{\mathbf{D}}$ are the ground-truth and predicted depth, respectively. $\mathbb{I}$ is the indicator function.}
    \label{tab:depthmetric}
\end{table}

For reconstruction metrics, we report Acc, Comp, Prec, Recall, F-score. Their definitions can be found in \cref{tab:reconmetric}.
We use a voxel size of 0.04m for TSDF reconstruction.
\begin{table}[ht]
    \centering
    \begin{tabular}{lc}
    \toprule
    Metric & Definition \\
    \midrule
    Acc & $\mbox{mean}_{p \in P}(\min_{p^*\in P^*}||p-p^*||)$ \\
    Comp & $\mbox{mean}_{p^* \in P^*}(\min_{p\in P}||p-p^*||)$ \\
    Prec & $\mbox{mean}_{p \in P}(\min_{p^*\in P^*}||p-p^*||<.05)$ \\
    Recal & $\mbox{mean}_{p^* \in P^*}(\min_{p\in P}||p-p^*||<.05)$ \\
    F-score & $\frac{ 2 \times \text{Perc} \times \text{Recal} }{\text{Prec} + \text{Recal}}$ \\
    \bottomrule
    \end{tabular}
    \caption{\textbf{Reconstruction metric definitions.} $P$ and $P^*$ are the point clouds sampled from predicted and ground truth mesh.}
    \label{tab:reconmetric}
\end{table}

\subsection{Baseline Details}
For the results presented in the main paper and supplementary video for Metric3D v2~\cite{hu2024metric3dv2} and Depth Pro~\cite{bochkovskii2024depth}, we input the ground-truth focal length into their models.
The ZoeDepth*~\cite{bhat2023zoedepth} model is trained using reproduced code from Depth Anything v1~\cite{yang2024depth}, and we utilize the base model of Depth Anything v1 for conducting experiments.
The MSPF results for ARKitScenes dataset are taken from~\cite{baruch2021arkitscenes}, and we retrain it using ScanNet++~\cite{yeshwanth2023scannet++} training data for testing on Scannet++ with the reproduced code from ARKitScenes.

\subsection{Ransac Alignment Details}
For monocular depth estimation methods, we perform a post-alignment to ensure fair comparison.
We utilize RANSAC alignment to align their output depth with the iPhone LiDAR depth.
Specifically, we first resize the output depth to match the dimensions of the iPhone LiDAR depth, then randomly formed several groups of samples. Each group of sample points is used to calculate a scale and shift, followed by voting using all points. 
The voting threshold is set to the median of the differences between the entire set of numbers and the median(Median Absolute Deviation). 
Then we apply the scale and shift to the predicted depth to align it with the ground-truth depth.
This method is more robust compared to the commonly used polyfit alignment in monocular depth estimation, typically improving the F-score by 8-10\% on ScanNet++ dataset.

\begin{figure}[htb]
    \centering 
    \includegraphics[width=0.99\columnwidth]{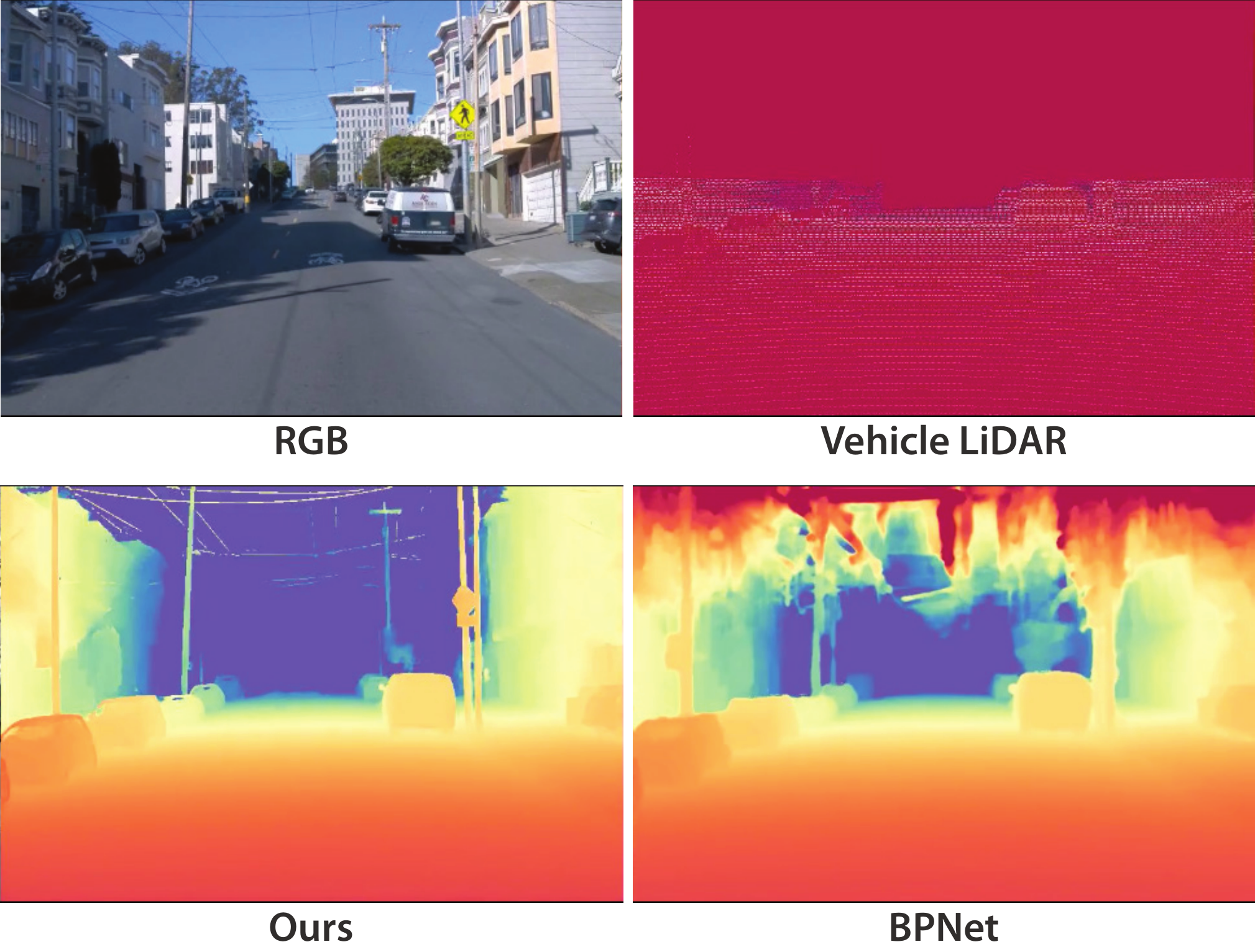}
    \caption{\textbf{Qualitative comparison of vehicle LiDAR completion.} We include more video results in the supplementary video. 
    }
    \label{fig:lidarcomp}
\end{figure}

\section{Prompting with a vehicle LiDAR}
We evaluate our method on the Waymo dataset to assess its performance with vehicle LiDAR.
Vehicle LiDAR significantly differs from the LiDAR used in smartphones, as it is generally coarse and consists of X-beam sparse LiDAR (typically 64 beams for Waymo dataset~\cite{sun2020scalability}).
Therefore, before inputting the data into the network, we perform KNN completion on the vehicle LiDAR depth ($k=4$).
We train our model on the Shift dataset~\cite{shift2022}, a synthetic dataset designed for autonomous driving, which includes RGB and depth data. The LiDAR data is simulated using the approach detailed in the main paper. We evaluate our model on the Waymo dataset.
We make comparisons with BPNet~\cite{BP-Net} in \cref{fig:lidarcomp}. Our method demonstrates precise depth estimation and we include more video results and street reconstruction results in the supplementary video.


\section{Generalized Robotic Grasping Details}

\paragraph{Detailed setups.}
We control the right arm of a Unitree H1 humanoid robot while fixing its lower body. The task is to grasp the object on the table and put it into the box, one at a time. 
The object is randomly placed at nearby, middle, and far positions.
The robot policy runs at 30 Hz. However, due to overheating issues in our lab environment, the iPhone can only stably capture images at 15 Hz, resulting in the visual input being updated every two control steps.


We first teleoperate the robot to collect 60, 80 trajectories for diffusive objects (red \& green cans) and transparent objects (glass bottles); then, we take the diffusive set of data as training set to train ACT~\cite{zhao2023learning} policies with different types of visual inputs, including the estimated depth by our model, 
ARKit depth directly from the iPhone, and also RGB images; during evaluation, we test the grasping performance corresponding to different visual inputs on all objects.

\paragraph{Model architectures.}
We use the same network structure with ACT~\cite{zhao2023learning} with one image input. ACT policy crops all types of visual input at 480x640 resolution and processes images with a pre-trained ResNet18 backbone\cite{he2016deep}. For depth images, the first layer of the pre-trained network is replaced with a 1-channel convolutional network.
The pretrained ResNet18 helps enhance the generalization of policy. Without the pretrained parameters, the policy with depth input only grasps the same position.

We include more video results in the supplementary video.

\end{document}